\PassOptionsToPackage{table}{xcolor}
\documentclass[11pt, letterpaper, logo]{googledeepmind}

\PassOptionsToPackage{comma,numbers,sort,compress}{natbib}

\usepackage[comma,numbers,sort,compress]{natbib}

\usepackage{hyperref}[citecolor=magenta]
\usepackage{fdsymbol}

\hypersetup{
    colorlinks = true,
    citecolor = {magenta},
}

\pdftrailerid{redacted}

\makeatletter
\renewcommand\bibentry[1]{\nocite{#1}{\frenchspacing\@nameuse{BR@r@#1\@extra@b@citeb}}}
\makeatother

\usepackage{kantlipsum, lipsum}
\usepackage{dsfont}
\usepackage{gdm-colors}

\usepackage{bbm}
\usepackage{wrapfig}
\usepackage{colortbl}
\usepackage{xcolor}
\usepackage[table]{xcolor}
\usepackage{tabularx}
\usepackage{graphicx}    % 用于加载图片
\usepackage{subcaption}  % 专门用于 \begin{subfigure} 环境
\usepackage{amsmath}     % 用于标题中的数学符号（如 \mathbf{P}）

\usepackage{caption}
\usepackage[percent]{overpic}
\usepackage{afterpage}
\usepackage{float}

\usepackage[most,skins,theorems]{tcolorbox}
\usepackage{orcidlink}
\usepackage{longtable}

\tcbset{
  aibox/.style={
    width=\linewidth,
    top=8pt,
    bottom=4pt,
    colback=blue!6!white,
    colframe=black,
    colbacktitle=black,
    enhanced,
    center,
    attach boxed title to top left={yshift=-0.1in,xshift=0.15in},
    boxed title style={boxrule=0pt,colframe=white,},
  }
}
\newtcolorbox{AIbox}[2][]{aibox,title=#2,#1}
\definecolor{lightblue}{rgb}{0.22,0.45,0.70}

\graphicspath{{photos/}}

\title{Quantitative Video World Model Evaluation for Geometric-Consistency}

\correspondingauthor{xueyanz@mail.tsinghua.edu.cn}

\author[1]{Jiaxin Wu}
\author[1]{Yihao Pi}
\author[2]{Yinling Zhang}
\author[3]{Yuheng Li}
\author[1]{Xueyan Zou}

\affil[1]{Tsinghua University - IEI Lab}
\affil[2]{UW-Madison}
\affil[3]{Adobe Research}
\vspace{-0.5cm}
\begin{abstract}
Generative video models are increasingly studied as implicit world models, yet evaluating whether they produce physically plausible 3D structure and motion remains challenging. 
Most existing video evaluation pipelines rely heavily on human judgment or learned graders, which can be subjective and weakly diagnostic for geometric failures. We introduce \textbf{PDI-Bench} (\textbf{P}erspective \textbf{D}istortion \textbf{I}ndex), a quantitative framework for auditing \emph{geometric coherence} in generated videos. Given a generated clip, we obtain object-centric observations via segmentation and point tracking (e.g., SAM 2, MegaSaM, and CoTracker3), lift them to 3D world-space coordinates via monocular reconstruction, and compute a set of projective-geometry residuals capturing three failure dimensions: \emph{scale--depth alignment}, \emph{3D motion consistency}, and \emph{3D structural rigidity}. To support systematic evaluation, we build \textbf{PDI-Dataset}, covering diverse scenarios designed to stress these geometric constraints. Across state-of-the-art video generators, PDI reveals consistent geometry-specific failure modes that are not captured by common perceptual metrics, and provides a diagnostic signal for progress toward physically grounded video generation and physical world model. Our code and dataset can be found at \url{https://pdi-bench.github.io/}.

\end{abstract}

\begin{document}
\maketitle

\enlargethispage{1.\baselineskip}

% 1. Talk about the recent advancement in video generation model, 

\begin{figure}[h]
    \centering
    \includegraphics[width=0.98\linewidth]{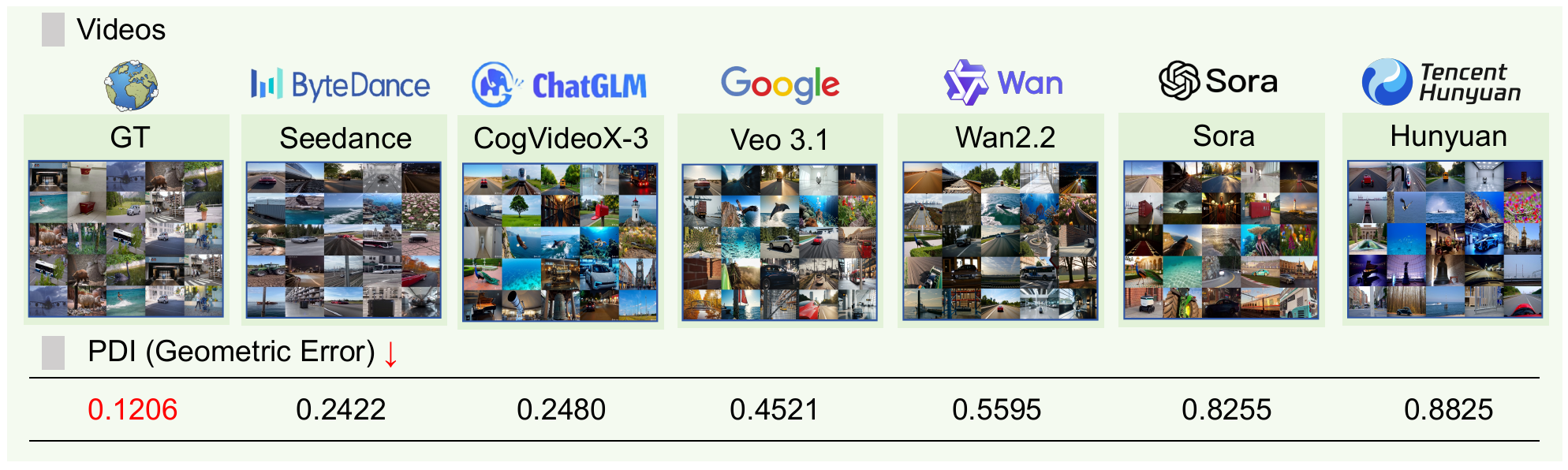}
    \vspace{-5pt}
    \caption{\textbf{Overview of the PDI-Bench Evaluation.} 
    \textbf{(Top)} Qualitative samples from our dataset, featuring real-world Ground Truth (GT) videos and generated sequences from state-of-the-art models. 
    \textbf{(Bottom)} The corresponding \textbf{PDI-Scores} for GT and each model. Lower scores indicate better adherence to 3D physical laws (scale alignment, motion consistency, and structural rigidity). 
    }
    \vspace{-10pt}
    \label{fig:pdi_overview}
\end{figure}

\section{Introduction}
% --- INTRODUCTION SECTION ---
\label{sec:intro}

% 1. What it has been achieved for video generation currently with a very impressive performance, and the improvement for video generation quality has already greatly improved for physical based consistency.
% The recent surge in high-fidelity generative video models, exemplified by Sora~\cite{liu2024sorareviewbackgroundtechnology}, Kling~\cite{klingteam2025klingomnitechnicalreport}, and Seedance~\cite{gao2025seedance10exploringboundaries}, has marked a paradigm shift in the creation of digital content. By leveraging massive spatio-temporal datasets, these models have achieved unprecedented textural realism and semantic alignment, leading many to categorize them as nascent ``World Models.'' This terminology implies that generative networks have moved beyond simple pixel interpolation toward a latent understanding of fundamental physical laws and the three-dimensional (3D) structure of our reality.
High-fidelity generative video models like Seedance~\cite{seedance2_2026}, Veo 3.1~\cite{google_flow_2024}, and Sora~\cite{liu2024sorareviewbackgroundtechnology, openai2024sora} have reshaped content creation with their unprecedented visual realism. This impressive quality has led many to view them as early "World Models". This terminology implies a critical shift from simple 2D pixel interpolation to a deeper, latent understanding of 3D structures and physical laws.

\vspace{15pt}
Despite their visual realism, a significant gap remains between visual plausibility and geometric rigor. Current state-of-the-art models frequently struggle with \textbf{spatial scale} and \textbf{perspective consistency}. While sometimes subtle, these artifacts clearly violate fundamental Euclidean properties. Common failure modes include ``volume breathing,'' where rigid objects unrealistically expand or contract, and ``skating,'' where an object's motion is decoupled from the ground plane's perspective. These geometric flaws stem from a lack of explicit structural constraints during generation, exacerbated by the absence of metrics capable of evaluating 3D properties from 2D videos. Existing metrics, such as Fréchet Video Distance (FVD)~\cite{unterthiner2019accurategenerativemodelsvideo} or CLIP-based scores~\cite{radford2021learningtransferablevisualmodels}, rely on pixel distributions or semantic features, rendering them inherently ``geometry-blind.'' They fail to penalize physical errors, such as a train shrinking disproportionately to its velocity. To progress toward true physical simulation, the field needs a new evaluation standard that assesses 2D video motion using the strict rules of 3D projective geometry.

\begin{figure}[t] % h:当前位置, t:页顶, b:页底, p:独立页
    \centering % 图片居中
    \includegraphics[width=1.0\textwidth]{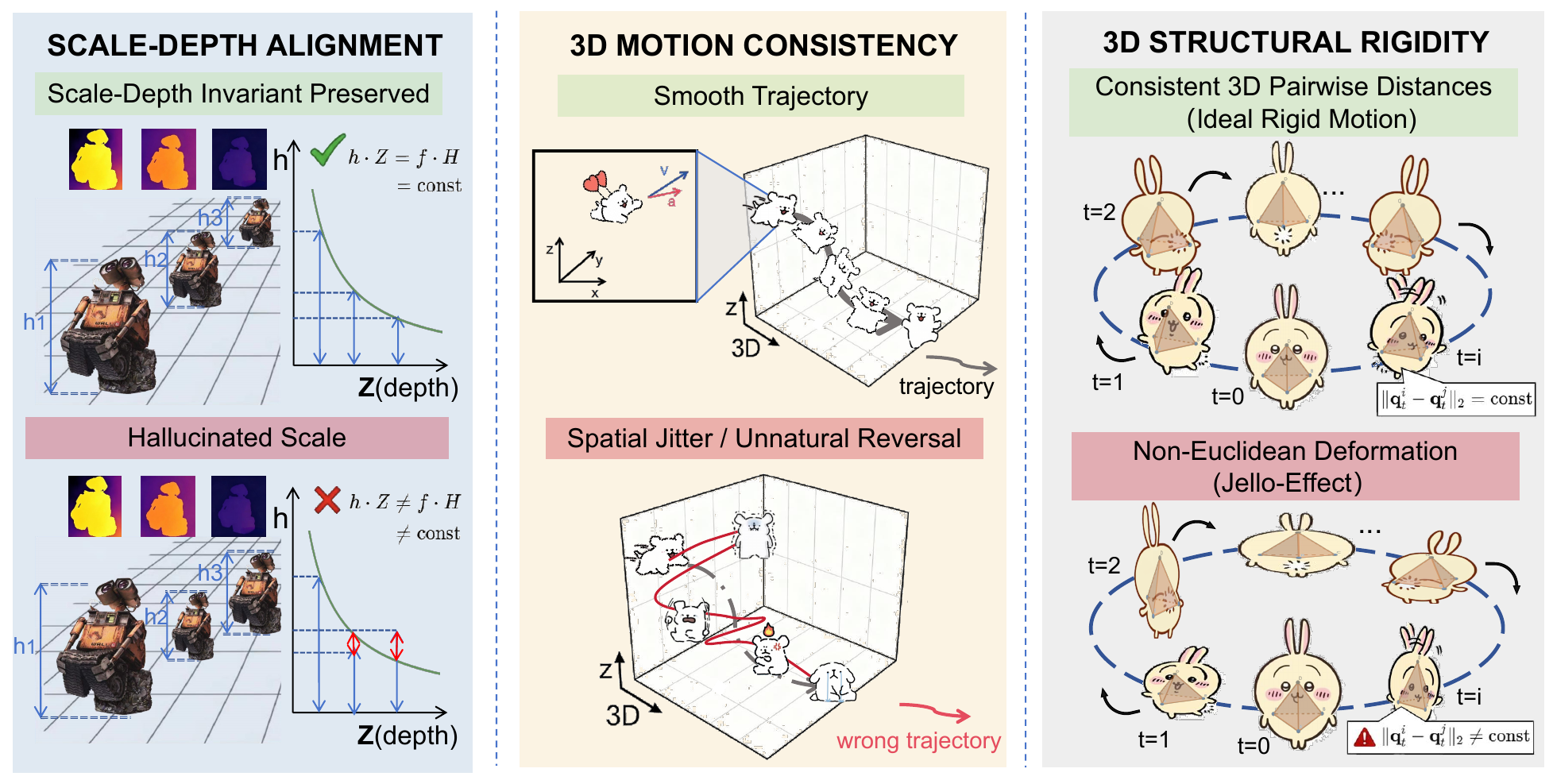} 
    \vspace{-15pt}
    \caption{\textbf{The Three key perspectives that PDI-Bench is evaluating for geometric consistency.} 
    \textbf{(Left) Scale-Depth Alignment:} Based on the pinhole camera model, we verify if the product of projected height $h$ and depth $Z$ remains constant. 
    \textbf{(Middle) 3D Motion Consistency:} We audit the smoothness of 3D world trajectories. 
    \textbf{(Right) 3D Structural Rigidity:} By monitoring internal 3D pairwise distances between anchors (skeleton $\mathbf{q}^i, \mathbf{q}^j$), we detect non-Euclidean deformations.} % 设置图片下方的标题
    \label{fig:metric} % 为图片设置标签，用于文中引用
    \vspace{-5pt}
\end{figure}

% The root cause of these geometric deficiencies lies in the lack of explicit structural constraints within the diffusion-based generation process and, perhaps more critically, the absence of quantitative evaluation frameworks capable of auditing 3D properties from monocular outputs. Existing metrics, such as Fréchet Video Distance (FVD)~\cite{unterthiner2019accurategenerativemodelsvideo} or CLIP-based scores~\cite{radford2021learningtransferablevisualmodels}, operate primarily on stochastic pixel distributions or high-level semantic embeddings. Consequently, they are inherently ``geometry-blind,'' failing to penalize physical inconsistencies such as a train shrinking at a rate inconsistent with its velocity, or a rigid object undergoing non-Euclidean deformation during curved motion. To advance generative video toward true physical simulation, the community requires a rigorous yardstick capable of translating 2D pixel dynamics into the 3D language of projective geometry.

While evaluating physical common sense has emerged as a key research frontier, existing suites remain insufficient for \textit{rigorous geometric verification}. Current pipelines typically depend on human judgment or rely on Large Multimodal Models (LMMs) as proxy evaluators, which are prone to subjective bias. Automated benchmarks like \textbf{PhysBench}~\cite{chow2025physbenchbenchmarkingenhancingvisionlanguage} and \textbf{WorldBench}~\cite{upadhyay2026worldbenchdisambiguatingphysicsdiagnostic} address scalability but primarily focus on high-level \textit{categorical} phenomena (e.g., gravity, buoyancy), leaving them largely ``geometry-agnostic.'' Conversely, emergent geometric metrics such as \textbf{MEt3R}~\cite{asim2026met3rmeasuringmultiviewconsistency} and \textbf{TRAJAN}~\cite{allen2025directmotionmodelsassessing} assess \textit{implicit consistency} via learned representations, but lack the transparency to diagnose specific physical violations. Our work, \textbf{PDI-Bench}, fills this critical gap by operationalizing \textit{explicit physical laws}—such as scale--depth alignment, 3D motion consistency, and 3D structural rigidity—as hard quantitative constraints. By lifting 2D video dynamics into verifiable world-space residuals, PDI-Bench provides a precise diagnostic signal to expose the underlying geometric hallucinations of generative world models.

% Although evaluation of physical world understanding has recently emerged as a critical frontier, existing diagnostic suites remain insufficient for rigorous physical auditing. Pioneering frameworks such as \textbf{PhysBench}~\cite{chow2025physbenchbenchmarkingenhancingvisionlanguage} and \textbf{WorldBench}~\cite{upadhyay2026worldbenchdisambiguatingphysicsdiagnostic} primarily focus on the recognition of high-level categorical phenomena, such as gravity and buoyancy. While these benchmarks excel at auditing semantic or causal logic, they remain largely ``geometry-agnostic.'' Specifically, they overlook the fine-grained, low-level geometric regularities---such as perspective scaling and vanishing point convergence---that form the fundamental Euclidean scaffolding of any physical simulation. Our work, \textbf{PDI-Bench}, complements these behavioral and causal assessments by providing a sub-pixel level audit of the geometric consistency inherent in world simulators.

As shown in Fig.~\ref{fig:metric}, \textbf{PDI-Bench} evaluates physical realism via three orthogonal geometric metrics. \textit{Scale-depth alignment} enforces the inverse correlation between an object's projected 2D height and its 3D depth to penalize unnatural scale hallucinations. \textit{3D motion consistency} evaluates the object's 3D centroid motion in world coordinates, penalizing non-physical accelerations and abrupt angular shifts (e.g., spatial jitter or teleports) while fully decoupling camera movements. \textit{3D structural rigidity} tracks internal 3D point-pair distances over time to penalize localized, non-Euclidean deformations (e.g., the ``jello effect'') and preserve rigid-body integrity.
% As shown in Fig.~\ref{fig:metric}, the proposed \textbf{PDI-Bench} evaluates physical realism by breaking down geometric consistency into four distinct, measurable metrics. It assesses \textit{scale-depth alignment} to ensure that an object's projected height correctly correlates with its depth, directly penalizing unnatural volume breathing. To evaluate motion plausibility, \textit{kinematic convergence} verifies that an object's displacement accurately synchronizes with its scaling rate relative to the scene's vanishing point. Furthermore, \textit{structural cohesion} tracks internal point-pair distances to penalize non-Euclidean, localized deformations, ensuring the object maintains rigid-body integrity. Lastly, \textbf{perspective coupling} audits the global spatial relationship by measuring the angular alignment between the foreground subject's motion and the background environment, exposing instances where objects appear to ``skate'' across the ground plane.

To move beyond subjective visual assessment, PDI-Bench extracts 3D geometric evidence from 2D pixels via a ``perception-to-reasoning'' \textbf{Target-Uplift-Anchor} workflow. First, \textbf{Semantic Targeting} via SAM~2~\cite{ravi2024sam2segmentimages} isolates the subject to establish precise 2D spatial boundaries and scale priors ($h$). Next, \textbf{3D Geometric Uplifting} via MegaSaM~\cite{li2024megasamaccuratefastrobust} reconstructs world-coordinate pointmaps and camera poses directly from the monocular sequence, lifting 2D observations into a unified 3D physical environment while fully decoupling camera ego-motion. Finally, \textbf{3D Structural Anchoring} via CoTracker3~\cite{karaev2024cotracker3simplerbetterpoint} deploys dense point anchors within the subject mask. By utilizing the 2D trajectories as precise spatial indices into the aforementioned 3D pointmaps, we lift visual cues into structurally meaningful 3D trajectories ($\mathbf{q}_t^n$), enabling camera-motion-invariant evaluation of both Motion consistency and structural rigidity.

% Moving beyond qualitative observation, PDI-Bench constructs a multi-dimensional ``perception-to-reasoning'' pipeline through a collaborative \textbf{Target-Uplift-Anchor} workflow: (i)~\textbf{Semantic Targeting} via SAM~2~\cite{ravi2024sam2segmentimages} to isolate auditing subjects and establish precise 2D scale priors ($h$); (ii)~\textbf{Long-range Tracking} utilizing CoTracker3~\cite{karaev2024cotracker3simplerbetterpoint} to deploy microscopic anchors within objects for monitoring motion vectors and structural stability; and (iii)~\textbf{3D Uplifting} employing Mega-SAM~\cite{li2024megasamaccuratefastrobust} to reconstruct latent depth ($Z$) trajectories and camera poses from monocular sequences.

% Central to our framework is the \textbf{Perspective Distortion Index (PDI)}, a comprehensive metric system derived from Euclidean projection axioms. PDI quantifies physical realism across four critical dimensions: \textit{Scale-Depth Alignment} ($\epsilon_{scale}$) verifying the $h \cdot Z$ conservation law; \textit{Kinematic Convergence Residual} ($\epsilon_{trajectory}$) auditing scaling-motion synchronization; \textit{Structural Cohesion} ($CV(R_{integrity})$) quantifying rigidity through point-pair distance analysis; and \textit{Perspective Coupling Consistency} ($\epsilon_{vp}$) measuring the angular alignment between foreground motion and background geometry.

\begin{figure}[t]
    \centering
    \includegraphics[width=0.98\linewidth]{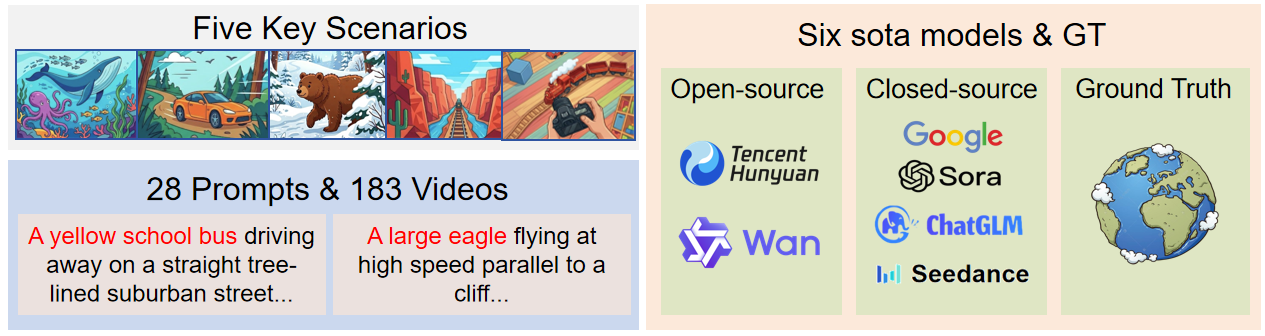}
    \vspace{-5pt}
    \caption{\textbf{Overview of the PDI-Dataset and Experimental Setup.} 
    Our benchmark comprises 183 high-quality videos generated from 28 diverse text prompts, evaluated against real-world Ground Truth (GT). We benchmark six state-of-the-art video generators, categorized into \textbf{Open-source} and \textbf{Closed-source} models . The dataset is meticulously curated to cover five critical physical scenarios:  (1) Longitudinal Convergence; (2) Dynamic Tracking; (3) Biological Motion; (4) Curved Motion; and (5) Partial Occlusion.}
    \label{fig:bench}
    \vspace{-5pt}
\end{figure}

% To facilitate a comprehensive evaluation, we contribute the \textbf{PDI-Dataset}, a diverse benchmark specifically curated to expose geometric vulnerabilities across six critical physical scenarios: (1)~\textit{Longitudinal Convergence} on linear paths; (2)~\textit{Dynamic Tracking} with moving backgrounds; (3)~\textit{Non-rigid Biological Motion}; (4)~\textit{Curved Motion}; (5)~\textit{Partial Occlusions}; and (6)~\textit{Hitchcock zoom} effects. Our dataset encompasses real-world Ground Truth (GT) and outputs from leading generative models, categorized into \textbf{Open-source} (e.g., Wan2.2~\cite{wanvideo2025wan22}, HunyuanVideo~\cite{kong2025hunyuanvideosystematicframeworklarge}) and \textbf{Closed-source} (e.g., CogVideoX~\cite{cogvideox2025web}, Sora~\cite{liu2024sorareviewbackgroundtechnology}, Flow~\cite{google_flow_2024}) tiers. Experimental results provide the first systematic quantification of spatial hallucinations in these models, establishing a rigorous geometric yardstick for assessing the physical intelligence of artificial world simulators.

In summary, we claim the following contributions:

\begin{itemize}[leftmargin=*, noitemsep, topsep=2pt, parsep=0pt, partopsep=0pt]
    \item We propose \textbf{PDI-Bench}, a quantitative framework that translates 2D pixel dynamics into 3D geometric reasoning to detect physical hallucinations.
    \item We introduce the \textbf{Perspective Distortion Index}, a metric system that utilizes 3D lifting to decouple kinematic errors from geometric alignment.
    \item We present \textbf{PDI-Dataset}, dedicated to geometric consistency, featuring real-world data and 6 SoTA open/closed-source video models.
    \item We demonstrate that PDI-Bench effectively identifies subtle spatial inconsistencies across five stress-test scenarios, offering critical insights for developing the next generation of space-aware generative systems.
    
\end{itemize}

\section{Related Work}
\label{sec:related}

\noindent
\textbf{Video Generation and World Model.} Recent breakthroughs in diffusion-based generative models have significantly advanced the state of video synthesis. 
Early works focused on temporal extensions of 2D diffusion processes~\cite{ho2020denoisingdiffusionprobabilisticmodels,blattmann2023stablevideodiffusionscaling}. 
More recently, large-scale models such as 
Sora~\cite{liu2024sorareviewbackgroundtechnology}, and Wan2.2~\cite{wanvideo2025wan22} have demonstrated
a remarkable ability to generate high-fidelity, long-duration sequences with complex semantic alignment.
The impressive visual quality has led to the conceptualization of these systems as ``World Models'', implying an internal representation of physical laws. However, whether these models truly simulate 3D space or simply replicate 2D statistical patterns remains a subject of intense debate.
Our work aims to provide a geometric yardstick to quantify this distinction.

\noindent
\textbf{Video Quality Assessment and Prompt Following.} The evaluation of generative videos has traditionally relied on distribution-based metrics such as Fréchet Video Distance (FVD)~\cite{unterthiner2019accurategenerativemodelsvideo} and Inception Score (IS)~\cite{salimans2016improvedtechniquestraininggans}, which prioritize frame-level textural and aesthetic quality over structural integrity. To assess text-video semantic alignment and prompt following, CLIP-based scores~\cite{radford2021learningtransferablevisualmodels} have become the standard. Building on these foundations, comprehensive suites like VBench~\cite{huang2023vbenchcomprehensivebenchmarksuite} and T2V-CompBench~\cite{sun2025t2vcompbenchcomprehensivebenchmarkcompositional} have recently been proposed to provide multi-dimensional evaluations, encompassing dynamic quality, temporal consistency, and compositional prompt adherence.

\noindent \textbf{Physical Consistency and Scene Understanding.} The transition from video synthesis to world simulation requires evaluating true physical realism rather than mere perceptual quality. While recent benchmarks assess macroscopic physical laws---such as VIDEOPHY~\cite{bansal2024videophyevaluatingphysicalcommonsense} and WorldModelBench~\cite{li2025worldmodelbenchjudgingvideogeneration} using VLM judges, PhysBench~\cite{chow2025physbenchbenchmarkingenhancingvisionlanguage} querying interaction dynamics, PhyGenBench~\cite{meng2024worldsimulatorcraftingphysical} testing dynamic constraints, and WorldBench~\cite{upadhyay2026worldbenchdisambiguatingphysicsdiagnostic} fitting physical constants---they primarily probe high-level semantic plausibility. More recently, \textbf{WorldScore}~\cite{duan2025worldscoreunifiedevaluationbenchmark} proposes a unified benchmark for world generation, focusing on long-range scene transitions and camera controllability via SLAM-based tracking and multi-modal assessment. 
Closer to our approach, \textbf{MEt3R}~\cite{asim2026met3rmeasuringmultiviewconsistency} assesses multi-view consistency by warping semantic features through latent 3D pointmaps, and \textbf{TRAJAN}~\cite{allen2025directmotionmodelsassessing} employs a trajectory autoencoder to detect anomalies in point tracks. However, these methods measure \textit{implicit consistency} via learned representations. PDI-Bench fills a critical gap by enforcing \textit{explicit physical laws} as quantitative constraints. Instead of evaluating high-level plausibility, our metric audits fundamental scale--depth alignment, 3D motion consistency, and structural rigidity to expose the spatial integrity of world models.

\begin{wrapfigure}{r}{0.45\textwidth} % {r}表示在右侧，{0.5\textwidth}表示宽度占页面的一半
    \centering
    \vspace{-7pt}
    \includegraphics[width=\linewidth]{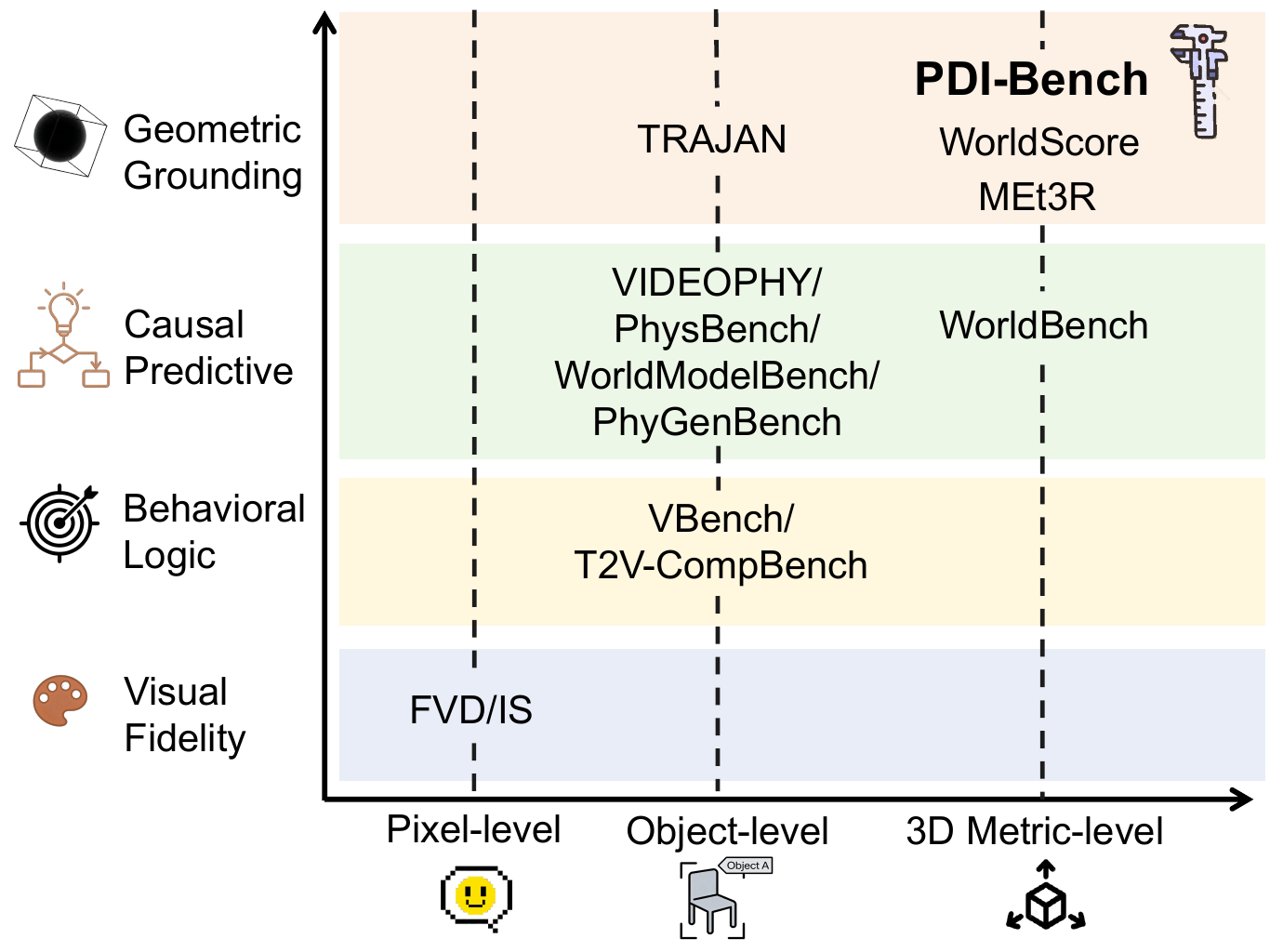}
    \label{fig:taxonomy}
\end{wrapfigure}
\noindent
\textbf{Visual Perception and 3D Reconstruction.} 
The feasibility of 3D physical auditing is underpinned by the surge in foundation models for visual perception. Video segmentation has been revolutionized by memory-based propagation models like SAM 2~\cite{ravi2024sam2segmentimages}, while temporal correspondence has reached pixel-level precision through dense trackers such as CoTracker3~\cite{karaev2024cotracker3simplerbetterpoint}. Furthermore, robust 3D reconstruction from dynamic monocular sequences has become accessible via semantic-aware SfM frameworks like MegaSaM~\cite{li2024megasamaccuratefastrobust}. These perceptual backbones serve as the foundational sensors in our pipeline, projecting 2D video dynamics into a verifiable 3D world coordinate system.

\section{Method}
The core objective of \textbf{PDI-Bench} is to bridge the gap between 2D pixel dynamics and 3D physical regularities. We operationalize this objective through a multi-stage \textit{Target-Uplift-Anchor} pipeline, culminating in a three-dimensional geometric audit quantified by our proposed \textbf{Perspective Distortion Index(PDI)}.

\subsection{Target-Uplift-Anchor: Perceptual Pipeline}
\label{pipeline}
\begin{figure}[t]
    \centering
    \includegraphics[width=0.98\linewidth]{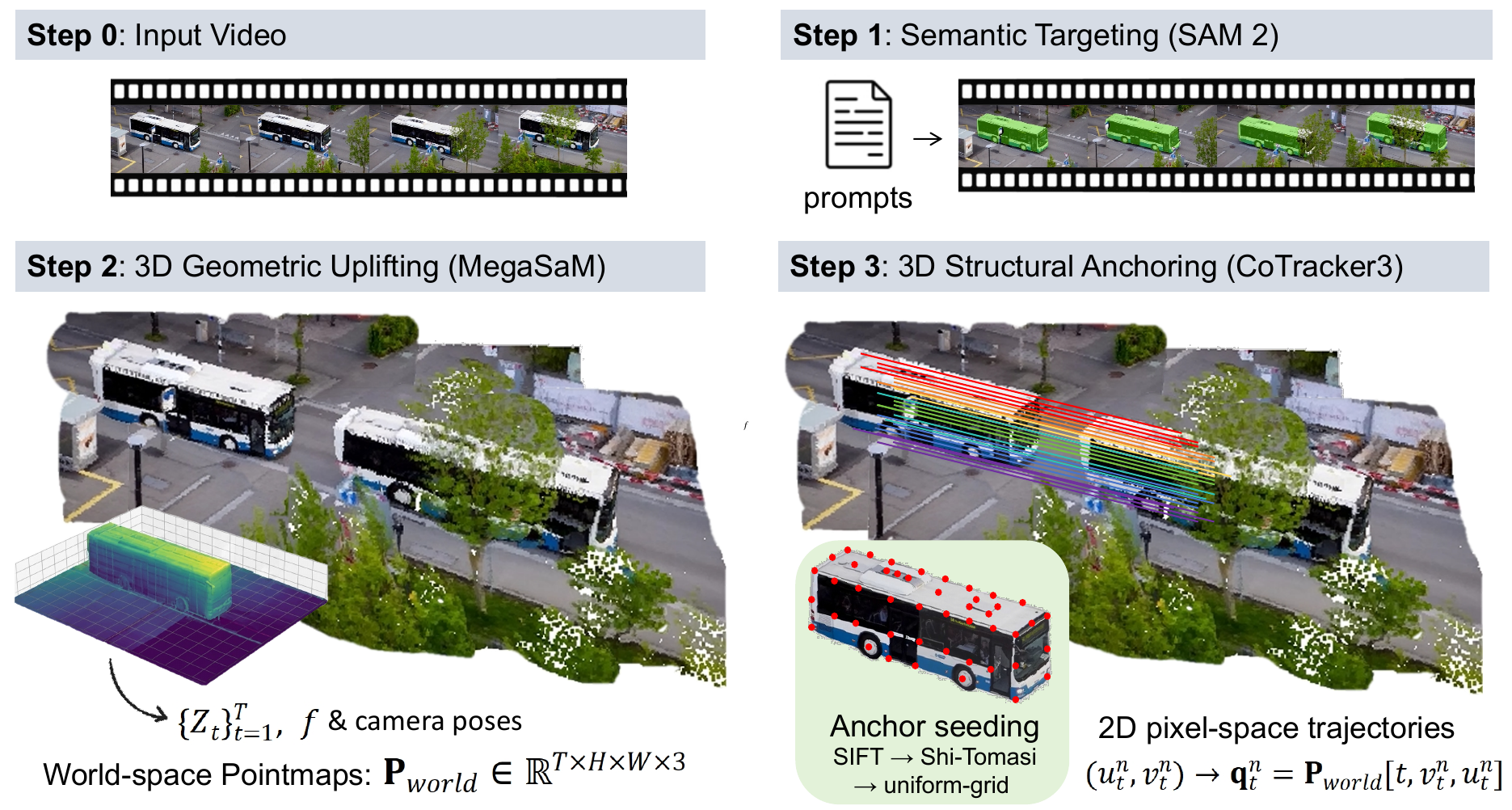}
    \vspace{-5pt}
    \caption{\textbf{Overview of the PDI-Bench Pipeline.} Our \textit{Target-Uplift-Anchor} pipeline initiates with \textbf{Semantic Targeting} (SAM~2) for object isolation. It then performs \textbf{Geometric Uplifting} via MegaSaM to construct 3D world-space pointmaps $\mathbf{P}_{world}$, followed by \textbf{Structural Anchoring} with CoTracker3 to lift 2D pixel-space trajectories into 3D coordinates $\mathbf{q}_t^n$. The resulting spatial-temporal data are synthesized into the PDI metric for consistency auditing. See Sec.~\ref{pipeline} for details.}
    \label{fig:PDI-pipeline}
    \vspace{-5pt}
\end{figure}

As shown in Fig.~\ref{fig:PDI-pipeline}, we orchestrate a hierarchical perception-to-reasoning stack through a collaborative \textit{Target-Uplift-Anchor} workflow to extract the physical evidence required for geometric auditing.

\textbf{Semantic Targeting (SAM 2).} We initiate the pipeline by identifying the auditing subject using \textbf{Florence-2}~\cite{xiao2023florence2advancingunifiedrepresentation} for automated text-to-box prompting. These prompts are then fed into \textbf{SAM~2}~\cite{ravi2024sam2segmentimages} to generate and propagate a temporal sequence of binary masks $\{M_t\}_{t=1}^T$. From these masks, we derive the instantaneous pixel height $h_t$ and the 2D spatial boundaries.

\textbf{3D Geometric Uplifting (MegaSaM).} To recover the latent 3D physical environment, we employ \textbf{MegaSaM}~\cite{li2024megasamaccuratefastrobust} to obtain a coherent depth sequence $\{Z_t\}_{t=1}^T$, the estimated focal length $f$, and camera poses. More importantly, MegaSaM projects every pixel into a unified 3D world coordinate system, yielding world-space pointmaps $\mathbf{P}_{world} \in \mathbb{R}^{T \times H \times W \times 3}$. This critical step lifts 2D observations into pure 3D space, completely decoupling object kinematics from camera ego-motion.

\textbf{3D Structural Anchoring (CoTracker3).} With the 3D world-space constructed, we deploy \textbf{CoTracker3}~\cite{karaev2024cotracker3simplerbetterpoint} to monitor the subject's internal structural integrity. Within the region defined by the initial mask $M_1$, we seed anchor queries via a SIFT $\to$ Shi-Tomasi $\to$ uniform-grid cascade. After filtering by visibility and displacement-jump thresholds, we obtain a set of reliable 2D pixel-space trajectories $\{(u_t^n, v_t^n)\}$. Using CoTracker3's 2D pixel-space trajectories $\{(u_t^n, v_t^n)\}$ as spatial indices into the MegaSaM world-space pointmaps $\mathbf{P}_{world}$, we lift each tracked anchor to its 3D coordinate: $\mathbf{q}_t^n = \mathbf{P}_{world}[t, v_t^n, u_t^n]$. This operation effectively transforms 2D visual tracking into structurally meaningful 3D trajectories for subsequent rigidity auditing.

Finally, the extracted multi-dimensional features from this \textit{Target-Uplift-Anchor} workflow are synthesized into the Perspective Distortion Index (PDI) via three orthogonal weighted residuals: $\epsilon_{scale}$, $\epsilon_{traj}$, and $\epsilon_{rigidity}$.

\subsection{The Perspective Distortion Index (PDI)}
\label{subsec:pdi}
We synthesize the multi-dimensional geometric evidence into the \textbf{Perspective Distortion Index (PDI)}, defined as a weighted sum of three orthogonal physical metrics:
\begin{equation}
  \text{PDI} = w_1 \cdot \operatorname{RMSE}(\epsilon_{scale}) + w_2 \cdot \operatorname{RMSE}(\epsilon_{traj}) + w_3 \cdot \epsilon_{rigidity}
\end{equation}
where $\sum_{i=1}^{3} w_i = 1$. Each term is designed to be scale-invariant and to capture a distinct failure mode. We apply the Root Mean Square Error (RMSE) to the scale and trajectory residuals to sensitize the index to catastrophic, high-magnitude physical hallucinations. Conversely, $\epsilon_{rigidity}$ is defined as the temporal mean of a robust dispersion statistic (MAD). We forgo RMSE for this term to avoid a redundant second-order penalty on an already aggregated measure of spatial incoherence, thereby preserving the direct physical interpretability of structural instability.

\subsubsection{Scale-Depth Alignment ($\epsilon_{scale}$)}

To establish a physical baseline, we model the generation process via pinhole camera geometry. For an object with physical height $H$ and depth $Z$, its projected pixel height $h$ satisfies $h = f \cdot H/Z$. Since $f$ and $H$ are constant for a rigid body, we derive the \textbf{Scale-Depth Invariant}:
\begin{equation}
  h_t \cdot Z_t = f \cdot H = \text{Constant}, \quad \text{where } h_t \in \text{SAM~2}, Z_t \in \text{MegaSaM}.
\end{equation}
Any fluctuation in this product indicates non-physical scaling (e.g., ``volume breathing''). We quantify this via a log-space residual $\epsilon_{scale}^{(t)}$ to ensure symmetric penalties for expansions and contractions:
\begin{equation}
\epsilon_{scale}^{(t)} = \left| \ln(h_t \cdot Z_t) - \operatorname{median}_{k \in [1,5]} \left(\ln(h_k \cdot Z_k)\right) \right|
\end{equation}
where the median of the first five frames establishes a stable baseline against initialization noise. The final $\text{RMSE}(\epsilon_{scale})$ serves as a robust measure of scaling severity, where $h_t$ and $Z_t$ are the per-frame pixel height and median object depth, respectively.

\subsubsection{3D Motion Consistency ($\epsilon_{traj}$)}

Rather than relying on 2D projective cues, we audit kinematic plausibility directly in the MegaSaM world coordinate system, which fully decouples object motion from camera ego-motion.

\textbf{Centroid Extraction \& Kinematics.} The per-frame 3D foreground centroid $\mathbf{C}_t$ is computed as the coordinate-wise median of all foreground points in the world-space pointmaps $\mathbf{P}_{world}$ masked by $M_t$. After applying temporal median filtering ($k\!=\!3$) to suppress depth flickering noise, we compute the frame-rate-normalized 3D velocity and acceleration using the frame interval $\Delta t = 1/\text{fps}$:
\begin{equation}
  \mathbf{v}_t = (\mathbf{C}_{t+1} - \mathbf{C}_t) / \Delta t, \qquad
  \mathbf{a}_t = (\mathbf{v}_{t+1} - \mathbf{v}_t) / \Delta t
\end{equation}

To evaluate adherence to Newtonian inertia, we decompose kinematic anomalies into two orthogonal, parallel components: abnormal acceleration magnitude and unnatural directional shifts.

\textbf{1. Acceleration Magnitude Penalty ($\tilde{a}_t$).} To quantify spatial jitter without bias from absolute scale, we define the relative acceleration ratio $r_t$ and its soft-saturated counterpart $\tilde{a}_t$ as:
\begin{equation}
  r_t = \frac{\|\mathbf{a}_t\|}{v_{\text{ref}}}, \quad 
  \tilde{a}_t = 2 \cdot \tanh(r_t / 5) \in [0, 2)
\end{equation}
where $v_{\text{ref}} = \max\!\left(\operatorname{median}(\|\mathbf{v}\|), 2\cdot\operatorname{median}(\|\mathbf{a}\|), \varepsilon\right)$ is a robust speed reference to prevent noise amplification near-stall ($\varepsilon=10^{-6}$). The $\tanh$ compression preserves linear scaling for minor deviations while bounding extreme outliers.

\textbf{2. Directional Continuity Penalty ($\varphi_t$).} Macroscopic objects cannot undergo instantaneous sharp turns without external forces. We penalize abrupt directional reversals via the cosine dissimilarity of consecutive velocity vectors. To prevent micro-tremors from triggering false penalties when the object is functionally stationary, this metric is activated only when the instantaneous speed exceeds a noise threshold ($\|\mathbf{v}\| > 0.1\,v_{\text{ref}}$):
\begin{equation}
  \varphi_t =
  \begin{cases}
      1 - \cos\angle(\mathbf{v}_{t-1},\,\mathbf{v}_t), & \text{if } \|\mathbf{v}_{t-1}\|, \|\mathbf{v}_t\| > 0.1\,v_{\text{ref}} \\
      0, & \text{otherwise}
  \end{cases}
\end{equation}
Note that $\varphi_t$ naturally falls within $[0, 2]$, where $0$ indicates identical directions and $2$ indicates a complete reversal, aligning perfectly with the scale of the magnitude penalty.

\textbf{Final Residual.} The magnitude and directional penalties are dimensionally aligned and combined with equal weights to form the holistic trajectory residual:
\begin{equation}
  \epsilon_{traj}^{(t)} = 0.5\cdot\tilde{a}_t + 0.5\cdot\varphi_t
\end{equation}

\subsubsection{Structural Rigidity ($\epsilon_{rigidity}$)}

To quantify non-physical internal deformations (e.g., the ``jello effect'' or ``volume breathing''), we audit the structural cohesion of the object across time based on 3D pairwise distance consistency. 

We obtain the 3D world-space coordinates $\mathbf{q}_t^n$ by sampling the MegaSaM pointmaps at the specific pixel locations defined by CoTracker3’s 2D trajectories. To guarantee high-fidelity correspondence and avoid boundary artifacts (such as depth bleeding), we select a set of optimal anchor pairs at the initial frame ($t\!=\!0$) through a rigorous triple-filtering strategy:
1) \textbf{Visibility:} Anchors must maintain a CoTracker3 tracking confidence $> 0.5$.
2) \textbf{Depth Smoothness:} We compute the Sobel gradient magnitude of the pointmap's Z-channel and discard points in the top-quartile ($> 75$th percentile) gradient regions to preclude depth discontinuities.
3) \textbf{Optimal Pair Scoring:} Valid points are paired by maximizing a joint heuristic score that balances the signal-to-noise ratio (large spatial separation) and inland reliability:$\mathcal{S}_{i,j} = \|\mathbf{q}_0^i - \mathbf{q}_0^j\|_2 \times \min(D_{mask}^i, D_{mask}^j)$, where $D_{mask}^i$ denotes the pixel distance from anchor $i$ to the nearest mask boundary.

In classical kinematics, a rigid body requires the 3D Euclidean distance between internal points to remain constant, i.e., $\|\mathbf{q}_t^i - \mathbf{q}_t^j\|_2 = \text{Constant}$. We define the distance ratio $r_{ij}(t)$ and the robust per-frame rigidity score as:
\begin{equation}
    r_{ij}(t) = \frac{\|\mathbf{q}_t^i - \mathbf{q}_t^j\|_2}{\|\mathbf{q}_0^i - \mathbf{q}_0^j\|_2}, \qquad
    \text{score}(t) = \frac{\operatorname{MAD}\bigl(\{r_{ij}(t)\}\bigr)}{\operatorname{median}\bigl(\{r_{ij}(t)\}\bigr) + \varepsilon}
\end{equation}
where MAD denotes the median absolute deviation, providing immunity against monocular global scale drift. Since $t\!=\!0$ inherently yields a zero score, it is excluded to prevent artificial suppression. The final rigidity metric is computed as the average over the active sequence:
\begin{equation}
  \epsilon_{rigidity} = \frac{1}{T-1}\sum_{t=1}^{T-1} \text{score}(t)
\end{equation}

% We define the \textit{Dissonance Ratio} $\mathcal{R} = \epsilon_{traj} / \epsilon_{scale}$, which measures the logical conflict between 2D kinematic observation and 3D metric uplifting. Since Mega-SAM decouples rotation from depth using multi-view constraints, $\epsilon_{scale}$ remains stable during rotation. If $\mathcal{R} > \tau$ (where $\tau=5$ is determined empirically), the system identifies a rotation-dominant case. In such instances, we adaptively refine the score: $\epsilon_{traj} \leftarrow \min(\epsilon_{traj}, \lambda \cdot \epsilon_{scale})$, where $\lambda=1.5$. This ensures PDI-Eval remains sensitive to AI-generated ``skating'' while maintaining robustness against complex, physically-grounded rotations.

\vspace{-10pt}
\section{Experiments}
\label{sec:exp}
In this section, we conduct a systematic evaluation of state-of-the-art generative video models using the \textbf{PDI-Bench} framework to quantify the discrepancy between visual plausibility and geometric consistency. By auditing diverse scenarios, we expose vulnerabilities in latent 3D spatial representations and provide a diagnostic roadmap for physically grounded world simulators.
\vspace{-3pt}
\subsection{Experimental Setup}
\label{sec:setup}
\textbf{PDI-Dataset and Evaluation Scenarios.} To systematically audit these generative systems, we curate a comprehensive benchmark containing 183 video sequences derived from 28 diverse textual prompts. This dataset comprises 15 high-quality, real-world Ground Truth (GT) videos serving as a reliable physical baseline for calibration, alongside 168 synthetic videos. To provide a holistic view of the field's current capabilities, we evaluate six representative state-of-the-art models categorized into two tiers: \textbf{Open-source} architectures (Wan 2.2~\cite{wanvideo2025wan22}, HunyuanVideo~\cite{kong2025hunyuanvideosystematicframeworklarge}) and \textbf{Closed-source} systems (Sora (OpenAI)~\cite{openai2024sora}, Seedance 2.0 Fast (ByteDance via Doubao)~\cite{seedance2_2026}, CogVideoX-3 (Zhipu AI via ChatGLM)~\cite{CogVideoX-32025web}, and Veo 3.1-Fast (Google via Flow)~\cite{google_flow_2024}). Furthermore, our evaluation is specifically designed to stress-test 3D spatial awareness by covering five critical geometric challenges: (1)~\textit{Longitudinal Convergence} (Longit. Conv.), (2)~\textit{Dynamic Tracking} (Dyn. Track.), (3)~\textit{Biological Motion} (Bio. Motion), (4)~\textit{Curved Motion} (Curved Mot.), and (5)~\textit{Partial Occlusion} (Part. Occl.). For the final PDI-score calculation, we empirically set the weights for scale-depth alignment, motion convergence, and structural rigidity to $(w_{scale}, w_{traj}, w_{rigidity}) = (0.4, 0.4, 0.2)$, respectively.

\subsection{Perceptual and Geometric Fidelity Guard}
To ensure the integrity of the PDI-Bench pipeline, we implement a multi-stage \textit{Fidelity Guard} that independently validates the outputs of the underlying perception models before PDI synthesis.

\textbf{Semantic Segmentation Audit (SAM 2):} 
To verify that the segmentation masks generated by SAM~2 consistently adhere to the intended target, we utilize a Vision-Language Model (VLM), specifically \textbf{Doubao}\cite{doubao2024}. We generate the RGB-mask overlays and the corresponding binary masks and prompt the VLM to evaluate whether the highlighted region accurately covers the object described by the \textit{text\_query}. 

\textbf{Point Tracking Audit (CoTracker3):}
We audit tracking quality using \textit{spatiotemporal trail maps} with cyan lines denoting historical trajectories. The VLM judges whether these trails represent the target object's motion trend. 

% \begin{figure}[b]
%   \centering
%   \includegraphics[width=0.98\linewidth]{photos/reconstruction1.pdf} % 这里的文件名后缀是 .pdf
%   \vspace{-10pt}
%   \caption{\textbf{VLM-based Fidelity Guard.} We employ a VLM to independently audit the semantic accuracy of SAM~2 masks (top) and the kinematic consistency of CoTracker3 trajectories (bottom).}
%   \label{fig:reconstruction1}
%   \vspace{-10pt}
% \end{figure}

\begin{figure}[t]
  \centering
  \includegraphics[width=0.98\linewidth]{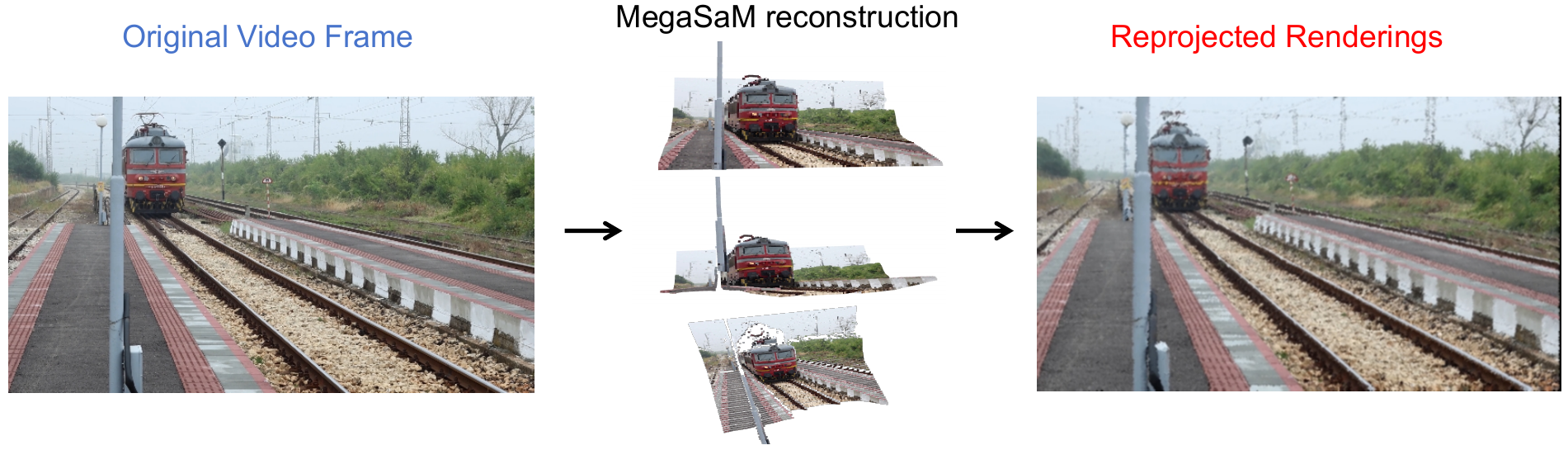} % 这里的文件名后缀是 .pdf
  \vspace{-5pt}
  \caption{\textbf{Reconstruction Audit.} We validate the fidelity of MegaSaM 3D pointmaps by reprojecting them onto target frames to ensure geometric consistency.}
  \label{fig:reconstruction2}
  \vspace{-5pt}
\end{figure}

\textbf{3D Reconstruction Audit (MegaSaM):} 
The fidelity of the 3D uplifted pointmaps is validated via a \textit{Cross-frame Reprojection Consistency} check. For a sampled pair of frames $\{I_A, I_B\}$, we re-synthesize the view of frame $B$ by projecting the world-space pointmaps of frame $A$ onto the image plane of $B$ using the estimated camera poses:
\begin{equation}
    \hat{I}_{B} = \mathcal{R}( \mathbf{P}_{world}^{A}, R_B, \mathbf{T}_B, K )
\end{equation}
where $\mathcal{R}$ denotes an off-screen renderer utilizing Z-buffering and point splatting. The reconstruction is deemed successful only if it satisfies predefined thresholds for \textit{Coverage} (density), \textit{MAE} (photometric error), and \textit{L2 distance}. A conceptual illustration of this validation scheme is provided in Fig.~\ref{fig:reconstruction2}.

\vspace{-5pt}
\subsection{Quantitative Evaluation via PDI}
\label{subsec:pdi_eval}
\vspace{-7pt}

Table~\ref{tab:pdi_results} presents a comprehensive ranking of state-of-the-art video generators based on our \textbf{PDI Score}. The results reveal a significant ``physics gap'' between perceived visual realism and underlying geometric consistency.

\textbf{Validation of Physical Baseline.} Real-world Ground Truth (GT) videos anchor the benchmark with a PDI Score of $0.1206$ and a remarkably low scale residual ($\epsilon_s = 0.0660$). This confirms that our 3D-uplifting pipeline accurately captures the near-perfect perspective laws of the physical world, setting a rigorous lower bound for generative auditing.

\textbf{Leading Generative Stability.} Among generative models, \textbf{Seedance 2.0} and \textbf{CogVideoX-3} exhibit the highest fidelity to physical laws. Seedance achieves the best overall stability with $0.0\%$ outliers and a leading \textit{MathPass} rate of $89.3\%$, suggesting superior geometric self-consistency. CogVideoX-3 demonstrates near-GT performance in \textit{Motion Consistency} ($\epsilon_t = 0.2033$) and \textit{Structural Rigidity} ($\epsilon_r = 0.2065$), indicating highly smooth and non-deformable object synthesis.

\textbf{The Scale Hallucination Crisis.} A critical finding is the high distortion observed in visually renowned models such as \textbf{Sora} and \textbf{HunyuanVideo}. Despite their aesthetic appeal, they suffer from severe \textit{scale hallucinations}, with $\epsilon_s$ values exceeding $1.67$ (a $25\times$ increase over GT). This highlights a fundamental failure in current transformer-based architectures to maintain the $h \cdot Z = \text{const}$ perspective invariant. Furthermore, their high standard deviations ($>1.7$) and outlier ratios ($14.3\%$) reflect a stochastic instability in physical modeling across different prompts.

\begin{table*}[t]
\centering
\vspace{-7pt}
\caption{\textbf{Quantitative Comparison of Physical Consistency on PDI-Bench.} We report the \textbf{PDI Score} (mean of residuals) and its breakdown. Lower values indicate higher physical realism. \textbf{GT} represents real-world videos as the benchmark baseline. Best generative results are \textbf{bolded}.}
\label{tab:pdi_results}
\resizebox{\linewidth}{!}{
\vspace{-10pt}
\begin{tabular}{llcccccccc}
\toprule
\textbf{Rank} & \textbf{Model} & \textbf{PDI Score} $\downarrow$ & \textbf{CI95} & \textbf{Scale} $\epsilon_{s} \downarrow$ & \textbf{Traj} $\epsilon_{t} \downarrow$ & \textbf{Rigid} $\epsilon_{r} \downarrow$ & \textbf{Std} & \textbf{Outlier} & \textbf{MathPass} $\uparrow$ \\
\midrule
1 & \textcolor{red}{Ground Truth (GT)} & \textcolor{red}{\textbf{0.1206}} & \textcolor{red}{[0.1018, 0.1386]} & \textcolor{red}{0.0660} & \textcolor{red}{0.1764} & \textcolor{red}{0.1182} & \textcolor{red}{0.0378} & \textcolor{red}{0.0\%} & \textcolor{red}{86.7\%}  \\
\midrule
2 & Seedance 2.0 & \textbf{0.2422} & [0.1954, 0.2920] & \textbf{0.2295} & 0.2064 & 0.3392 & 0.1315 & \textbf{0.0\%} & \textbf{89.3\%} \\
3 & CogVideoX-3 & 0.2480 & [0.1656, 0.4093] & 0.3135 & \textbf{0.2033} & \textbf{0.2065} & 0.3065 & 3.6\% & 85.7\% \\
4 & Veo 3.1 & 0.4521 & [0.2611, 0.7247] & 0.7507 & 0.2271 & 0.3049 & 0.6980 & 7.1\% & 50.0\% \\
5 & Wan 2.2 & 0.5595 & [0.2572, 1.0766] & 0.9317 & 0.2096 & 0.5150 & 1.2301 & 7.1\% & 67.9\% \\
6 & Sora & 0.8255 & [0.2652, 1.4847] & 1.6753 & 0.2711 & 0.2345 & 1.7312 & 14.3\% & 70.4\% \\
7 & HunyuanVideo & 0.8825 & [0.3094, 1.6018] & 1.8469 & 0.2515 & 0.2160 & 1.7730 & 14.3\% & 57.1\% \\
\bottomrule
\end{tabular}
}

\end{table*}

\vspace{-5pt}

\subsection{Category-Specific Physical Consistency Analysis}
\label{subsec:category_analysis}

Building on Table~\ref{tab:pdi_subtables_all} (a)--(e), we analyze how different motion patterns trigger specific physical hallucinations.

\textbf{Longitudinal Convergence.} This scenario audits scaling during axial motion. \textbf{HunyuanVideo} ($PDI=0.10$) and \textbf{CogVideoX-3} ($PDI=0.15$) approach the GT baseline, demonstrating a strong grasp of perspective laws. Conversely, \textbf{Wan2.2} and \textbf{Veo 3.1} exhibit high scale errors ($\epsilon_s > 0.32$), resulting in a ``sliding'' effect where object size mismatches relative 3D depth.

\textbf{Biological Motion.} Evaluating articulated dynamics reveals the difficulty of maintaining consistency in non-rigid entities. \textbf{Seedance 2.0} leads this category ($PDI=0.25$), effectively preserving structural integrity. In contrast, \textbf{Veo 3.1} and \textbf{HunyuanVideo} suffer from ``volumetric breathing'' hallucinations, where body mass fluctuates inconsistently ($\epsilon_s > 1.97$) during gait cycles.

\textbf{Curved Motion.} Non-linear trajectories and rotations represent the most significant challenge. \textbf{Sora} exhibits catastrophic failure here ($PDI=2.13$), driven by massive scale distortion ($\epsilon_s = 4.87$). This suggests that transformer-based generators often fail to preserve the $h \cdot Z$ invariant during rotational transformations. \textbf{CogVideoX-3} and \textbf{Seedance 2.0} remain the most robust in this scenario.

\textbf{Partial Occlusion.} This tests spatial memory and object permanence. \textbf{HunyuanVideo} shows severe degradation ($PDI=2.41, \epsilon_s = 5.38$), indicating the model ``forgets'' physical dimensions while the object is obscured. Conversely, \textbf{Sora} and \textbf{Seedance 2.0} demonstrate superior resilience, maintaining structural cohesion ($\epsilon_r < 0.45$) upon the object's re-emergence.

\textbf{Dynamic Tracking.} This scenario assesses the decoupling of camera ego-motion from object kinematics. \textbf{CogVideoX-3} ($PDI=0.16$) and \textbf{HunyuanVideo} ($PDI=0.17$) excel, nearing GT fidelity. However, \textbf{Sora} struggles significantly with scale ($\epsilon_s = 2.84$), as its world model tends to conflate camera proximity with non-physical object growth, leading to a collapse of perspective coupling.

\begin{table*}[t]
\centering
\scriptsize % 统一字号

% ==========================================
% 第一排：物理场景 (a), (b), (c)
% ==========================================
\begin{minipage}[t]{0.32\textwidth}
    \centering
    \textbf{(a) Longitudinal Convergence} \\
    \resizebox{\linewidth}{!}{
    \begin{tabular}{lcccc}
        \toprule
        Model & PDI $\downarrow$ & Scale $\downarrow$ & Traj $\downarrow$ & Rigid $\downarrow$ \\
        \midrule
        \rowcolor{white} \textbf{\color{red}{GT (ref)}} & \textbf{\color{red}{0.0715}} & \textbf{\color{red}{0.0198}} & \textbf{\color{red}{0.1406}} & \textbf{\color{red}{0.0369}} \\
        HunyuanVideo  & 0.1004 & 0.0580 & 0.1533 & 0.0795 \\
        CogVideoX-3 & 0.1540 & 0.0814 & 0.1909 & 0.2252 \\
        Sora & 0.2496 & 0.2656 & 0.2634 & 0.1903 \\
        Seedance 2.0 & 0.2623 & 0.2084 & 0.2351 & 0.4245 \\
        Veo 3.1 & 0.2958 & 0.4344 & 0.2115 & 0.1869 \\
        Wan2.2 & 0.3305 & 0.3278 & 0.2273 & 0.5424 \\
        \bottomrule
    \end{tabular}}
\end{minipage}
\hfill
\begin{minipage}[t]{0.32\textwidth}
    \centering
    \textbf{(b) Dynamic Tracking} \\
    \resizebox{\linewidth}{!}{
    \begin{tabular}{lcccc}
        \toprule
        Model & PDI $\downarrow$ & Scale $\downarrow$ & Traj $\downarrow$ & Rigid $\downarrow$ \\
        \midrule
        \rowcolor{white} \textbf{\color{red}{GT (ref)}} & \textbf{\color{red}{0.1167}} & \textbf{\color{red}{0.0623}} & \textbf{\color{red}{0.1734}} & \textbf{\color{red}{0.1123}} \\
        CogVideoX-3 & 0.1620 & 0.1039 & 0.2222 & 0.1577 \\
        HunyuanVideo  & 0.1722 & 0.1839 & 0.2097 & 0.0738 \\
        Veo 3.1 & 0.2170 & 0.3233 & 0.1677 & 0.1029 \\
        Seedance 2.0 & 0.2334 & 0.2214 & 0.2039 & 0.3162 \\
        Wan2.2 & 0.2941 & 0.1661 & 0.2141 & 0.7098 \\
        Sora & 1.2824 & 2.8387 & 0.2719 & 0.1910 \\
        \bottomrule
    \end{tabular}}
\end{minipage}
\hfill
\begin{minipage}[t]{0.32\textwidth}
    \centering
    \textbf{(c) Biological Motion} \\
    \resizebox{\linewidth}{!}{
    \begin{tabular}{lcccc}
        \toprule
        Model & PDI $\downarrow$ & Scale $\downarrow$ & Traj $\downarrow$ & Rigid $\downarrow$ \\
        \midrule
        \rowcolor{white} \textbf{\color{red}{GT (ref)}} & \textbf{\color{red}{0.1319}} & \textbf{\color{red}{0.0887}} & \textbf{\color{red}{0.1950}} & \textbf{\color{red}{0.0921}} \\
        Seedance 2.0 & 0.2536 & 0.3401 & 0.1816 & 0.2246 \\
        CogVideoX-3 & 0.2773 & 0.4031 & 0.1781 & 0.2239 \\
        Wan2.2 & 0.2827 & 0.3632 & 0.1748 & 0.3377 \\
        Sora & 0.3924 & 0.5968 & 0.2555 & 0.2571 \\
        HunyuanVideo  & 0.9760 & 2.1265 & 0.2310 & 0.1649 \\
        Veo 3.1 & 1.0230 & 1.9738 & 0.2268 & 0.7135 \\
        \bottomrule
    \end{tabular}}
\end{minipage}

\vspace{1.5em} % 增加垂直间距，防止上下排挤在一起

% ==========================================
% 第二排：物理场景 (d), (e) 和 独立的人类研究表
% ==========================================
\begin{minipage}[t]{0.32\textwidth}
    \centering
    \textbf{(d) Curved Motion} \\
    \resizebox{\linewidth}{!}{
    \begin{tabular}{lcccc}
        \toprule
        Model & PDI $\downarrow$ & Scale $\downarrow$ & Traj $\downarrow$ & Rigid $\downarrow$ \\
        \midrule
        \rowcolor{white} \textbf{\color{red}{GT (ref)}} & \textbf{\color{red}{0.1550}} & \textbf{\color{red}{0.1136}} & \textbf{\color{red}{0.1366}} & \textbf{\color{red}{0.2745}} \\
        Seedance 2.0 & 0.2558 & 0.3001 & 0.2122 & 0.2542 \\
        CogVideoX-3 & 0.2567 & 0.2944 & 0.2174 & 0.2600 \\
        Wan2.2 & 0.5223 & 0.8862 & 0.2042 & 0.4305 \\
        Veo 3.1 & 0.6037 & 1.0484 & 0.2728 & 0.3759 \\
        HunyuanVideo  & 0.7467 & 1.4705 & 0.2321 & 0.3282 \\
        Sora & 2.1277 & 4.8660 & 0.3225 & 0.2617 \\
        \bottomrule
    \end{tabular}}
\end{minipage}
\hfill
\begin{minipage}[t]{0.32\textwidth}
    \centering
    \textbf{(e) Partial Occlusion} \\
    \resizebox{\linewidth}{!}{
    \begin{tabular}{lcccc}
        \toprule
        Model & PDI $\downarrow$ & Scale $\downarrow$ & Traj $\downarrow$ & Rigid $\downarrow$ \\
        \midrule
        \rowcolor{white} \textbf{\color{red}{GT (ref)}} & \textbf{\color{red}{0.1346}} & \textbf{\color{red}{0.0626}} & \textbf{\color{red}{0.2115}} & \textbf{\color{red}{0.1250}} \\
        Seedance 2.0 & 0.2101 & 0.1075 & 0.1960 & 0.4433 \\
        Sora & 0.2201 & 0.1617 & 0.2482 & 0.2807 \\
        Veo 3.1 & 0.2414 & 0.2270 & 0.2640 & 0.2251 \\
        CogVideoX-3 & 0.3964 & 0.6964 & 0.2058 & 0.1774 \\
        Wan2.2 & 1.3157 & 2.8131 & 0.2208 & 0.5109 \\
        HunyuanVideo  & 2.4104 & 5.3793 & 0.4248 & 0.4436 \\
        \bottomrule
    \end{tabular}}
\end{minipage}
\hfill
\begin{minipage}[t]{0.32\textwidth}
    \centering
    \textbf{(f) Human Expert Study} \\
    \resizebox{\linewidth}{!}{
    \begin{tabular}{clcc}
        \toprule
        Rank & Model & Mean Score $\downarrow$ & Std. Dev. \\
        \midrule
        1 & \textbf{\color{red}{Ground Truth (GT)}} & \textbf{1.57} & 1.00 \\
        2 & Seedance 2.0 & 2.96 & 1.86 \\
        3 & CogVideoX-3 & 2.97 & 1.47 \\
        4 & Veo 3.1 & 3.36 & 1.56 \\
        5 & Wan2.2 & 3.37 & 2.26 \\
        6 & Sora & 3.60 & 1.99 \\
        7 & HunyuanVideo & 3.66 & 2.29 \\
        \bottomrule
    \end{tabular}}
\end{minipage}

\caption{
\textbf{Component-wise Physical Consistency and Human Alignment.}
(a)--(e) \textbf{Component-wise Error Analysis across Different Scenarios.} Each sub-table evaluates models on a specific challenge. GT (ref) denotes the real-world baseline. Lower values indicate higher physical consistency.
(f) \textbf{Human Expert Evaluation Results.} Scores range from 1 (Best) to 10 (Worst). The human consensus ranking exhibits a perfect alignment ($\rho=1.0$) with our automated PDI Score.
}
\vspace{-13pt}
\label{tab:pdi_subtables_all}
\end{table*}

\subsection{Human Expert Study and Perceptual Alignment}
\label{subsec:human_study}
To validate the alignment between \textbf{PDI-Bench} and human physical intuition, we conducted a perceptual study involving seven computer vision experts. The evaluation set comprised 105 unique video clips: 15 real-world (GT) videos and 90 AI-generated clips (15 per model) sourced from six state-of-the-art generators. For each model, three clips were sampled for each of the five physical scenarios. Adhering to the PDI scoring protocol, experts assigned ratings ranging from \textbf{1 (Physical Realism)} to \textbf{10 (Catastrophic Failure)}. With each clip reviewed by all seven experts, we obtained $n=105$ subjective ratings per model.

\textbf{Results and Alignment Analysis.} As summarized in Table~\ref{tab:pdi_subtables_all} (f),  the expert consensus yields a model ranking \textbf{identical} to our automated results. Real-world GT consistently anchors the high-fidelity bound with the lowest mean score ($1.57 \pm 1.00$), confirming the experts' ability to distinguish natural physics from synthetic artifacts. Among the AI models, \textit{Seedance 2.0} and \textit{CogVideoX-3} emerge as the top performers with nearly identical scores (2.96 and 2.97, respectively). Their low mean scores suggest a higher degree of temporal consistency and fewer visible violations of basic physical laws. \textit{Veo 3.1} and \textit{Wan 2.2} occupy the middle tier (3.36--3.37). While competitive, they exhibit more frequent physical inaccuracies compared to the top-tier models. In contrast, \textit{Sora} and \textit{HunyuanVideo} were penalised for the severe ``scale hallucinations'' and ``motion jitter'' previously identified by our 3D geometric audit. Notably, the high standard deviations in the lower-ranked models (e.g., $2.29$ for HunyuanVideo) suggest that their failure modes are often dramatic and highly visible to human observers. This consistent alignment underscores the validity of PDI-Bench as an objective, automated proxy for human perceptual assessment of physical laws in video generation.

\vspace{-5pt}
\section{Case Study: Diagnosing Autoregressive Extrapolation}
\label{subsec:case_study_ar}

Beyond standard evaluation, we conducted a stress test on autoregressive (AR) long-video generation. We evaluate the \textbf{Self-Forcing} paradigm~\cite{huang2025selfforcingbridgingtraintest} built on the Wan2.1-T2V-1.3B architecture~\cite{wan2025wanopenadvancedlargescale}. By extrapolating 81-frame trained sequences to 129 frames across 28 prompts (see the Appendix~\ref{app:prompt_gallery}), we analyze the decay of 3D physical consistency beyond the training context window.

\begin{table}[h]
    \centering
    \small % 使用标准表格字号，比正文略小，更显专业
    \setlength{\tabcolsep}{3pt} % 进一步紧缩列间距
    
    % 调整比例：给表格 0.52 宽度，给文字 0.45 宽度
    \begin{minipage}[c]{0.52\linewidth} 
        \centering
        \captionof{table}{\textbf{Self-Forcing AR Analysis.} Scale drift occurs despite stable trajectories.}
        \label{tab:self_forcing_ar}
        \vspace{-5pt}
        \begin{tabular*}{\linewidth}{@{\extracolsep{\fill}}lcccc}
            \toprule
            \textbf{Category} & \textbf{PDI} $\downarrow$ & \textbf{Scale} $\downarrow$ & \textbf{Traj} $\downarrow$ & \textbf{Rigid} $\downarrow$ \\
            \midrule
            Longit. Conv.  & 1.3063 & 2.8462 & 0.3078 & 0.2237 \\
            Dynamic Track. & 0.1994 & 0.1515 & 0.2604 & 0.1730 \\
            Biological Mot. & 2.0819 & 4.6326 & 0.3489 & 0.4464 \\
            Curved Motion   & 0.3111 & 0.3158 & 0.2530 & 0.4177 \\
            Partial Occl.  & 2.7570 & 6.2172 & 0.4097 & 0.5311 \\
            \midrule
            \textbf{Overall Mean} & \textbf{1.3407} & \textbf{2.8583} & \textbf{0.3170} & \textbf{0.3531} \\
            \bottomrule
        \end{tabular*}
    \end{minipage}
    \hfill
    \begin{minipage}[c]{0.45\linewidth} 
        % 右侧文字也同步使用 \small 或保持默认
        \normalsize 
        \textbf{Kinematic Success vs. Geometric Collapse.} Table~\ref{tab:self_forcing_ar} reveals a striking dichotomy in the AR model's physical capabilities. Across all scenarios, the 3D Kinematic Trajectory error ($\epsilon_{t}$) remains remarkably stable (overall mean: $0.3170$). This indicates that the Self-Forcing training paradigm, combined with rolling KV caching, successfully mitigates high-frequency spatial jitter and unnatural reversals often seen in long AR generation. 

    \end{minipage}
\end{table}
\vspace{-5pt}
% To demonstrate the diagnostic power of PDI-Bench beyond standard evaluation, we conduct a stress test on autoregressive (AR) long-video generation. We evaluate the recently proposed \textbf{Self-Forcing}~\cite{huang2025selfforcingbridgingtraintest} paradigm, built upon the Wan2.1-T2V-1.3B~\cite{wan2025wanopenadvancedlargescale} architecture. Although natively trained on 81-frame sequences, we extrapolate the model to 129 frames to observe the decay of 3D physical consistency as generation exceeds the training context window. This analysis is conducted across the 28 prompts detailed in Appendix~\ref{app:prompt_gallery}, capturing how geometric coherence degrades during long-horizon autoregressive rollout.

However, this kinematic smoothness masks a catastrophic failure in 3D projective geometry. The overall Scale-Depth Alignment error ($\epsilon_{s}$) surges to $2.8583$, particularly in \textit{Longitudinal Convergence} and \textit{Biological Motion}. This demonstrates severe \textit{scale hallucination}: as the model extrapolates beyond its 81-frame training horizon, it loses the ``spatial memory'' of the object's original 3D volume, causing objects to expand or contract independently of their depth variations.

\textbf{Vulnerability to Spatial Disconnects.} The data further reveal that AR generation is highly sensitive to continuous visual context. In \textit{Dynamic Tracking}, where the subject remains centrally focused, the scale error remains low ($0.1515$). Conversely, in \textit{Partial Occlusion}, the model exhibits its worst performance ($PDI=2.7570$). When the object is temporarily obscured, the AR mechanism loses its structural anchor within the context window, causing a complete failure in object permanence when the subject re-emerges. In \textit{Curved Motion}, while trajectory is maintained, the complexity of rotation leads to a higher rigidity residual ($\epsilon_{r}=0.4177$), indicating a mild ``jello effect'' during non-linear displacement. These insights highlight PDI-Bench's unique capability to provide fine-grained, frame-level diagnostic signals for emerging video generation architectures.

\section{Conclusion}
\label{sec:conclusion}

In this paper, we introduced \textbf{PDI-Bench}, a novel quantitative framework for auditing the perspective and scale consistency of generative video world models. By constructing a collaborative \textit{Target-Uplift-Anchor} workflow, we successfully translated 2D pixel dynamics into verifiable 3D geometric reasoning. Our proposed \textbf{Perspective Distortion Index (PDI)} provides a multi-modal metric to identify subtle hallucinations such as ``volume breathing'' and ``skating'' effects. Through extensive benchmarking on our \textbf{PDI-Dataset}, we demonstrated that our framework offers a robust geometric yardstick that complements existing semantic-based metrics. We believe PDI-Bench will serve as a foundational tool for evaluating and improving the physical intelligence of future artificial world simulators.

\noindent \textbf{Limitations.} While PDI-Bench establishes a rigorous geometric yardstick, it presents three primary limitations. First, its accuracy relies on off-the-shelf perception tools; in degraded videos where 3D uplifting fails, the framework must fall back to 2D proxies, reducing depth-aware precision. Second, our geometric invariants rely on a rigid-body assumption, making the metric less theoretically suited for highly non-rigid or amorphous subjects. Finally, disentangling complex 3D rotation from axial translation using purely monocular cues is fundamentally ill-posed, occasionally introducing measurement noise despite our consensus-based mitigations.

% Bibliography components
\bibliographystyle{abbrvnat}
\nobibliography*
\bibliography{googledeepmind-test}

\newpage
\appendix

\section{Additional Experimental Details}
\label{app:exp_details}

\subsection{PDI-Dataset Construction}
\label{app:dataset}
The PDI-Dataset consists of 183 video sequences in total, partitioned into real-world and synthetic subsets. 

\paragraph{Real-world sequences.}
The real-world portion of PDI-Dataset contains 15 short clips collected from the public video platform Pexels (\url{https://www.pexels.com/}).
We manually search for casual handheld or gimbal-assisted footage in ordinary environments and select clips that mirror the five core geometric scenarios introduced in the main paper: Longitudinal Convergence, Dynamic Tracking, Biological Motion, Curved Motion and Partial Occlusion.
All videos are manually inspected to ensure (i)~clear visibility of the auditing subject, (ii)~sufficient parallax for 3D reconstruction, and (iii)~minimal motion blur or rolling-shutter artifacts.

\paragraph{Synthetic sequences.}
The synthetic portion comprises 168 videos generated from 28 text prompts, each instantiated once by six representative generative video models: Wan~2.2\cite{wanvideo2025wan22}, HunyuanVideo\cite{kong2025hunyuanvideosystematicframeworklarge}, CogVideoX-3\cite{CogVideoX-32025web}, Seedance 2.0-Fast\cite{seedance2_2026}, Sora\cite{liu2024sorareviewbackgroundtechnology}, and Veo 3.1-Fast \cite{google_flow_2024}. 

\paragraph{Note on Model Variants.} 
To maintain transparency and reproducibility, we clarify the sources of the generative models evaluated in this benchmark. The \textbf{CogVideoX-3} results reported herein were obtained via the official \textit{Zhipu Qingying} web platform (accessed via ChatGLM). It is important to note that this represents a high-performance proprietary variant and should not be conflated with the open-weights \textit{CogVideoX-5B} or other community-driven iterations. Similarly, \textbf{Seedance 2.0-Fast}, \textbf{Sora} (OpenAI) and \textbf{Veo 3.1-Fast} (Google via Flow) were accessed via their respective commercial web interfaces as of April 2026. All synthetic videos presented in our benchmark reflect the baseline commercial performance available to end-users at the time of evaluation. Note that the Sora samples in our dataset were generated using the \$20 monthly consumer subscription rather than the enterprise API, representing the baseline commercial performance of the model.

The 28 text prompts are grouped into the five stress-test scenarios listed above and correspond to the high-level descriptions used in our PDI-Dataset (e.g., a car on a straight road for Longitudinal Convergence, a bus entering a roundabout or a tractor turning at a field boundary for Curved Motion).
For each model, we follow its official inference configuration (e.g., guidance scale, number of diffusion steps, native resolution) and, when the interface allows, fix the random seed to reduce stochastic variance; otherwise, we always use the first clip returned by the model to avoid cherry-picking.

\paragraph{Temporal resolution and pre-processing.}
All clips are normalized to a fixed frame rate of 24\,fps and trimmed to a duration of 4--12 seconds, leading to 96--300 frames per video.
For models that natively output longer videos, we uniformly crop a central window to avoid initialization artifacts and abrupt endings.
Each frame is resized to a resolution of $H \times W = 512 \times 512$ pixels while preserving the aspect ratio via zero-padding if necessary.

\subsection{Complete PDI-Bench Prompt Gallery}
\label{app:prompt_gallery}

To ensure the reproducibility of our benchmark and facilitate future comparisons, we provide the complete list of text prompts used to generate the video sequences in PDI-Bench. As detailed in Table~\ref{tab:all_prompts}, the dataset comprises 28 meticulously engineered prompts distributed across five distinct projective geometric scenarios: Longitudinal Convergence, Dynamic Tracking, Biological Motion, Curved Motion, and Partial Occlusion. Each prompt is designed to isolate specific 3D-to-2D spatial transformations and camera-object interactions.

{ % 使用大括号限制 arraystretch 的作用范围
\renewcommand{\arraystretch}{1.2}
\begin{longtable}{@{} l p{0.88\textwidth} @{}}
\caption{\textbf{Complete List of Prompts in PDI-Bench.} The prompts are carefully designed to evaluate specific geometric transformations and camera motions across six challenging categories.}
\label{tab:all_prompts} \\
\toprule
\textbf{Category} & \textbf{Text Prompt} \\
\midrule
\endfirsthead

% --- 这是跨页后的新表头 ---
\multicolumn{2}{@{}l}{\tablename~\thetable\ -- \textit{Continued from previous page}} \\
\toprule
\textbf{Category} & \textbf{Text Prompt} \\
\midrule
\endhead

% --- 这是跨页前的表尾提示 ---
\midrule
\multicolumn{2}{r@{}}{\textit{Continued on next page}} \\
\endfoot

% --- 这是表格最后一页的真正底部 ---
\bottomrule
\endlastfoot

% ================= 表格内容开始 =================

% --- Category 1 ---
\rowcolor{gray!15} \multicolumn{2}{@{}l}{\textbf{Longitudinal Convergence}} \\
& 1. A handheld following shot of a red vintage car driving away on a straight desert highway, harsh noon light and heat haze on the horizon, subtle shake and lateral drift. \\
& 2. A high-speed train moving toward the viewer on a straight track, low-angle handheld perspective, rails and gravel receding toward a clear vanishing point. \\
& 3. A yellow school bus driving away on a straight tree-lined suburban street, the shot tracking from a low position behind, morning light and clean asphalt. \\
& 4. A silver metallic sphere rolling away on a long reflective marble floor in a bright gallery, the shot following closely with slight sway. \\
& 5. A heavy cargo truck moving away on a straight bridge at night, tail lights glowing, subtle frame shake, city lights in the distance. \\
& 6. A large shipping container being pushed away on a straight industrial dock, cranes and water behind, moving viewpoint, overcast industrial light. \\
\midrule

% --- Category 2 ---
\rowcolor{gray!15} \multicolumn{2}{@{}l}{\textbf{Dynamic Tracking}} \\
& 1. A handheld following shot of a red sports car driving on a straight multi-lane highway, city skyline and roadside trees in the background receding rapidly with parallax. \\
& 2. A smooth following shot of an autonomous suitcase moving through a vast airport terminal, repeated columns and floor patterns rushing past in frame. \\
& 3. A close handheld shot following a large chrome sphere rolling along a straight, reflective museum corridor, exhibits and windows flowing past. \\
& 4. A following shot from a vehicle alongside, keeping pace with a large truck carrying a blue container on a long bridge, waves and bridge cables creating dynamic background motion. \\
& 5. A smooth following shot of a metal logistics crate moving along a straight automated conveyor, complex factory machinery in the background rushing past. \\
& 6. A handheld following shot of a large metal ball rolling through a straight modern art gallery, surrounding artworks and viewers receding rapidly with parallax. \\
\midrule

% --- Category 3 ---
\rowcolor{gray!15} \multicolumn{2}{@{}l}{\textbf{Biological Motion}} \\
& 1. A smooth following shot of a large eagle flying at high speed parallel to a cliff, rock face and sea below, clear sky. \\
& 2. A following shot from a moving boat of a dolphin swimming and leaping in the waves alongside, spray and sunlight. \\
& 3. A handheld shot of a large octopus swimming away in a complex coral reef, tentacles waving, colorful fish and coral, blue water and light shafts. \\
& 4. A backward-moving shot following a snake slithering through dense colorful flowers on the ground, petals and stems, soft daylight. \\
& 5. A moving shot following a peacock walking and shaking its tail feathers in a palace garden, fountains and trimmed hedges, ornate tiles. \\
\midrule

% --- Category 4 ---
\rowcolor{gray!15} \multicolumn{2}{@{}l}{\textbf{Curved Motion}} \\
& 1. A handheld tracking perspective follows a silver compact SUV navigating a sharp hairpin turn on a winding mountain road. The view orbits slightly to capture the vehicle transitioning from a front-view to a side-view against the pine forest background. \\
& 2. A low-angle shot follows a sports car drifting through a 90-degree corner on a professional race track. The car rotates intensely while the moving shot emphasizes the shifting vanishing lines of the curb and tire marks. \\
& 3. A cinematic tracking shot follows a city bus driving through a large, ornate stone roundabout. The view maintains a side perspective, showing the bus constantly changing its orientation relative to the central fountain and surrounding city traffic. \\
& 4. A ground-level perspective tracking a small delivery robot as it makes a sharp turn at a sidewalk corner. The shot stays close, highlighting the rotation of the robot's boxy frame against the detailed brickwork. \\
& 5. A handheld shot follows a green tractor making a wide turn at the edge of a plowed field. The view moves with the vehicle, capturing the shifting angles of the heavy wheels and mechanical parts against the vast landscape. \\
\midrule

% --- Category 5 ---
\rowcolor{gray!15} \multicolumn{2}{@{}l}{\textbf{Partial Occlusion}} \\
& 1. A car driving along a street at night, wheels briefly obscured by a low roadside guardrail for under a second, handheld shot moving alongside, street lamps and storefronts. \\
& 2. A train passing behind a row of thin vertical power line poles, the shot tracking its movement from a moving platform, sky and industrial landscape. \\
& 3. A bus moving through a city street, briefly partially hidden by a thin traffic sign, the shot following from the sidewalk. \\
& 4. A vintage car driving past a row of thin trees, never fully leaving the moving view, autumn leaves and road. \\
& 5. A boat sailing behind a thin pier support, remaining partially visible throughout, handheld shot from the dock, sea and sky. \\
& 6. A robot crate moving through a warehouse, passing behind a thin metal rack, the shot following alongside, shelves and boxes, industrial lighting. \\
\end{longtable}
}

\paragraph{Reconstruction-aware weighting.}
The final PDI score is synthesized as a weighted sum of three orthogonal physical residuals:
\begin{equation}
    \text{PDI Score} = w_1 \cdot \text{RMSE}(\epsilon_{scale}) + w_2 \cdot \text{RMSE}(\epsilon_{traj}) + w_3 \cdot \epsilon_{rigidity} ,
\end{equation}
where $\sum_i w_i = 1$. When the 3D reconstruction from MegaSAM passes our quality check (i.e., satisfying the ground-plane SVD and reprojection constraints), we adopt a uniform prior with $(w_1, w_2, w_3) = (0.4, 0.4, 0.2)$, assigning nearly equal importance to spatial scaling, temporal kinematics, and structural integrity.

\subsection{Evaluation Protocol}

For each model and scenario, we compute PDI scores on all valid sequences and report the \textbf{median} and the \textbf{95\% Bootstrap Confidence Interval (CI)} to ensure robustness against generative outliers. To guarantee statistical significance, each video is evaluated multiple times with different random seeds for anchor sampling, and the resulting scores are averaged. 

To facilitate cross-model comparison, we employ a \textbf{GT-Anchored Normalization} scheme. We first calculate robust statistics (Median and MAD) for each residual dimension across the real-world Ground Truth (GT) subset to define the "physics-perfect" baseline. Each raw residual is then standardized into a robust Z-score and mapped to a $[0, 100]$ score using a scaled half-logistic function. This dual-track reporting—providing both raw physical residuals (PDI-Error) and normalized scores (PDI-Score)—allows for a transparent and interpretable assessment of the physical common sense embedded in state-of-the-art video generators.

\section{Computing Resources}
\label{sec:compute_resources}

All experiments and evaluations in this work were conducted on a Linux workstation equipped with NVIDIA RTX 3090 GPUs. The detailed hardware is as follows:

\begin{itemize}
    \item \textbf{Hardware Configuration:}
    \begin{itemize}
        \item \textbf{GPU:} NVIDIA GeForce RTX 3090.
        \item \textbf{Video Memory:} 24,576 MiB (24 GB) GDDR6X per GPU.
        \item \textbf{Compute Capability:} 8.6 (Ampere architecture).
    \end{itemize}
    
    \item \textbf{Resource Utilization:} The PDI-Bench pipeline leverages multi-GPU parallelization to process the benchmark. Each video sequence is typically assigned to a single GPU worker to perform semantic segmentation (SAM2), point tracking (Co-Tracker), and 3D reconstruction (Mega-SAM) sequentially.
\end{itemize}

\bigskip

\section{Proofs of Main Theorems}
\label{app:proofs}

In this section, we provide formal derivations for the geometric identities and metric properties that underlie the Perspective Distortion Index (PDI).

\subsection{Geometric Invariants of Perspective Projection}

Let a 3D point $\mathbf{P} = (X, Y, Z)$ be projected onto the image plane at $\mathbf{p} = (x, y)$ with focal length $f$.
By similar triangles, we have
\begin{equation}
    \frac{x}{f} = \frac{X}{Z}, \quad \frac{y}{f} = \frac{Y}{Z}.
\end{equation}

Consider an object of physical height $H$ aligned vertically in 3D.
Let $(X, Y_1, Z)$ and $(X, Y_2, Z)$ denote its bottom and top endpoints.
The projected pixel height $h$ is given by $h = |y_2 - y_1|$.
Using the relation above,
\begin{equation}
    h = \left|\frac{f Y_2}{Z} - \frac{f Y_1}{Z}\right| = \frac{f |Y_2 - Y_1|}{Z} = \frac{f H}{Z}.
\end{equation}
Similarly, the horizontal coordinate of the object's centroid satisfies $x = f X / Z$.
Under a rigid-body assumption ($H$ and $X$ constant), we obtain the parameter-free invariant
\begin{equation}
    \frac{h}{x} = \frac{f H / Z}{f X / Z} = \frac{H}{X},
\end{equation}
establishing that the ratio of projected height to radial displacement is constant for any linear motion.

\subsection{Square-Inverse Scaling Law}

Starting from $h = f H / Z$, treat $H$ and $f$ as constants and differentiate with respect to $Z$:
\begin{equation}
    \frac{d h}{d Z} = - f H Z^{-2} = - \frac{f H}{Z^2}.
\end{equation}
For two nearby depths $Z_1$ and $Z_2 = Z_1 + \Delta Z$, a first-order Taylor expansion gives
\begin{equation}
    \Delta h \approx \frac{d h}{d Z} \bigg|_{Z = \bar{Z}} \cdot \Delta Z
    = - \frac{f H}{\bar{Z}^2} \Delta Z,
\end{equation}
where $\bar{Z}$ lies between $Z_1$ and $Z_2$.
Using the identity $\bar{Z}^2 \approx Z_1 Z_2$ for small relative changes, we obtain the square-inverse law reported in the main text:
\begin{equation}
    \Delta h \approx - \frac{f H \, \Delta Z}{Z_1 Z_2}.
\end{equation}
This shows that physically correct scaling must obey a quadratic suppression with depth; any linear-in-$1/Z$ behavior corresponds to a systematic velocity distortion.

\subsection{Spatio-temporal Coupling in the Image Plane}

We next derive the depth-coupled suppression of pixel-wise motion used in Eq.~(3) of the main text.
Starting from the pinhole relation for the horizontal coordinate,
\begin{equation}
    x = f_x \frac{X}{Z},
\end{equation}
consider a 3D displacement $(\Delta X, \Delta Y, \Delta Z)$ that moves the point from $(X, Y, Z)$ to $(X + \Delta X, Y + \Delta Y, Z + \Delta Z)$.
The new image coordinate is
\begin{equation}
    x' = f_x \frac{X + \Delta X}{Z + \Delta Z}.
\end{equation}
The induced pixel displacement is therefore
\begin{align}
    \Delta x
    &= x' - x
     = f_x \left( \frac{X + \Delta X}{Z + \Delta Z} - \frac{X}{Z} \right) \nonumber \\
    &= f_x \frac{Z(X + \Delta X) - X(Z + \Delta Z)}{Z(Z + \Delta Z)} \nonumber \\
    &= f_x \frac{Z \Delta X - X \Delta Z}{Z(Z + \Delta Z)}.
\end{align}
This recovers the expression used in the main paper and shows that pixel motion is quadratically suppressed with depth for fixed 3D velocities.

\subsection{Metric Implementation Details}

\paragraph{Scale-depth residual (Scale).}
Given per-frame object pixel height $h_t$ (from SAM2 masks) and aligned depth $z_t$ (from Mega-SAM), we audit the perspective-scale invariant in log space:
\[
s_t=\ln(\max(h_t,\epsilon))+\ln(\max(z_t,\epsilon)).
\]
Using the first $n_{\mathrm{ref}}=\min(5,T)$ frames, we set
\[
s_{\mathrm{ref}}=\mathrm{median}(s_1,\ldots,s_{n_{\mathrm{ref}}}),
\]
and compute framewise residuals
\[
\epsilon_{\mathrm{scale}}^{(t)}=\left|s_t-s_{\mathrm{ref}}\right|,\quad t\ge 2.
\]
The Scale component entering PDI is
\[
E_{\mathrm{scale}}=\mathrm{RMSE}\!\left(\epsilon_{\mathrm{scale}}\right).
\]
This term directly measures violation of the $h_tz_t\approx\mathrm{const}$ relation under perspective projection.

\paragraph{Motion consistency (Traj).}
We use a 3D kinematic audit in world coordinates from Mega-SAM pointmaps.
For each frame, we extract a robust foreground centroid by masked 3D median pooling; if valid foreground points are insufficient, we inherit the last valid centroid.
The centroid trajectory is then temporally denoised by a 1D median filter ($k=3$) per coordinate.
With $\Delta t=1/\mathrm{fps}$, velocity and acceleration are
\[
\mathbf{v}_t=\frac{\mathbf{x}_{t+1}-\mathbf{x}_t}{\Delta t},\qquad
\mathbf{a}_t=\frac{\mathbf{v}_{t+1}-\mathbf{v}_t}{\Delta t}.
\]
We define a robust speed reference
\[
s_{\mathrm{ref}}=\max\!\big(\mathrm{median}(\|\mathbf{v}\|),\,2\cdot \mathrm{median}(\|\mathbf{a}\|),\,10^{-6}\big),
\]
and acceleration penalty
\[
p_a^{(t)}=2\tanh\!\left(\frac{\|\mathbf{a}_t\|/s_{\mathrm{ref}}}{5}\right).
\]
Direction-change penalty is
\[
p_\theta^{(t)}=
\begin{cases}
1-\cos\theta_t, & \text{if both adjacent speeds are sufficiently large},\\
0, & \text{otherwise},
\end{cases}
\]
where $\theta_t$ is the angle between adjacent velocity vectors.
The trajectory residual is
\[
\epsilon_{\mathrm{traj}}^{(t)}=\tfrac12 p_a^{(t)}+\tfrac12 p_\theta^{(t)},
\]
and the Traj component is
\[
E_{\mathrm{traj}}=\mathrm{RMSE}\!\left(\epsilon_{\mathrm{traj}}\right).
\]

\paragraph{Structural Rigidity (Rigidity).}
To quantify non-physical internal deformation (e.g., the ``jello effect''), we evaluate object structural stability using a prioritized three-strategy hierarchy, and the active strategy output is directly used as the third PDI component.

\textbf{1) 3D Pairwise Rigidity (Primary).}
We sample world-space points $\mathbf{q}_t^n$ from Mega-SAM pointmaps at CoTracker locations. Anchor pairs are selected at $t=0$ by triple filtering: (i) visibility filtering, (ii) depth-gradient reliability filtering, and (iii) pair scoring that favors both large 3D separation and interior-region reliability (distance to mask boundary). For each frame,
\[
r_{ij}(t)=\frac{\|\mathbf{q}_t^i-\mathbf{q}_t^j\|_2}{\|\mathbf{q}_0^i-\mathbf{q}_0^j\|_2},
\quad
\rho_t=\frac{\mathrm{MAD}(\{r_{ij}(t)\})}{\mathrm{median}(\{r_{ij}(t)\})+\epsilon}.
\]
The strategy-1 rigidity residual is the temporal mean over $t\ge 2$:
\[
\epsilon_{\mathrm{rigid}}^{(1)}=\frac{1}{T-1}\sum_{t=2}^{T}\rho_t.
\]
(Frames with insufficient visible pairs inherit the previous frame score.)

\textbf{2) 3D Height Stability (Fallback when Strategy 1 is not entered).}
If 3D points are valid but strategy 1 is unavailable at the dispatcher level, we compute per-frame 3D object height from foreground $y$-span:
\[
h_t^{3D}=P_{95}(y_t)-P_{5}(y_t),
\]
and use coefficient of variation:
\[
\epsilon_{\mathrm{rigid}}^{(2)}=
\frac{\mathrm{std}(\{h_t^{3D}\}_{t=1}^{T})}
{\mathrm{mean}(\{h_t^{3D}\}_{t=1}^{T})+\epsilon}.
\]

\textbf{3) 2D Pairwise Consistency (Degraded fallback).}
When 3D evidence is unavailable, we use 2D CoTracker pairwise distance ratios:
\[
r_{ij}^{2D}(t)=\frac{d_{ij}(t)}{d_{ij}(0)},\qquad
\rho_t^{2D}=\frac{\mathrm{std}(\{r_{ij}^{2D}(t)\})}{\mathrm{mean}(\{r_{ij}^{2D}(t)\})+\epsilon},
\]
and compute
\[
\epsilon_{\mathrm{rigid}}^{(3)}=\frac{1}{T}\sum_{t=1}^{T}\rho_t^{2D}.
\]

Finally, the rigidity component used by PDI is
\[
\epsilon_{\mathrm{rigid}}=
\begin{cases}
\epsilon_{\mathrm{rigid}}^{(1)}, & \text{if Strategy 1 is selected},\\
\epsilon_{\mathrm{rigid}}^{(2)}, & \text{else if Strategy 2 is selected},\\
\epsilon_{\mathrm{rigid}}^{(3)}, & \text{else if Strategy 3 is selected},\\
0, & \text{if no strategy is available}.
\end{cases}
\]

\section{Geometric Invariants and Perspective Coupling in Longitudinal Motion}
\subsection{Diagnostic Perspective Analysis for Longitudinal Convergence}
\label{subsec:hvp_main}

For scenarios involving longitudinal or oblique motion (e.g., objects receding along a path), we introduce the \textbf{Generalized H-VP Homogeneity} as an auxiliary diagnostic tool. This constraint evaluates whether the projected centroid of a rigid object converges to the scene's vanishing point ($VP$) in synchronization with its scale reduction. Under the pinhole camera model, this yields a coupled inverse-depth law:
\begin{equation}
\frac{h_1}{h_t} = \frac{Z_t}{Z_1} = \frac{\operatorname{Dist}(\mathbf{p}_1, VP)}{\operatorname{Dist}(\mathbf{p}_t, VP)},
\label{eq:hvp}
\end{equation}
where $h_t$ is the pixel height, $Z_t$ is the 3D depth, and $\mathbf{p}_t$ is the image centroid at frame $t$. 

While our primary PDI-Score focuses on metrics applicable to arbitrary motion, Eq.~\eqref{eq:hvp} provides a deeper geometric probe specifically for the \textit{Longitudinal Convergence} category. It ensures that the object's ``speed'' of receding (trajectory) and its ``rate'' of shrinking (scale) are physically coupled. In practice, this distance-based formulation is used when a stable $VP$ can be estimated; for near-transverse motion, we transition to the angular formulation described below.

\subsection{Perspective Coupling via Angular Alignment}
\label{subsec:vp_alignment}

To further quantify the coupling between foreground motion and background environment (detecting the ``sticker-on-screen'' effect), we calculate the angular divergence between two independent vanishing points: the motion-derived $VP_{fg}$ and the geometry-derived $VP_{bg}$ (extracted via LSD line clustering). Let $(c_x, c_y)$ be the principal point; we define the respective direction vectors as:
\[
\mathbf{d}_{fg} = VP_{fg} - (c_x, c_y), \qquad \mathbf{d}_{bg} = VP_{bg} - (c_x, c_y).
\]
The perspective coupling residual, denoted as $\Delta_{\theta}$, is computed via cosine similarity:
\begin{equation}
\Delta_{\theta} = \frac{1-\cos\angle(\mathbf{d}_{fg}, \mathbf{d}_{bg})}{2},
\end{equation}
where $\Delta_{\theta} \in [0, 1]$. A value of $0$ indicates that the object is moving perfectly toward the scene's natural horizon, while $1$ indicates a total contradiction in perspective spaces. We utilize this angular metric instead of Euclidean distance to remain robust against transverse motions where $VP$ coordinates tend toward infinity.

\subsection{Mathematical Derivation of H-VP Homogeneity}
\label{subsec:hvp_derivation}
\begin{figure}[t]
    \centering
    % --- 左边：Camera (子图 a) ---
    \begin{subfigure}[b]{0.30\linewidth}
        \centering
        \includegraphics[width=\linewidth]{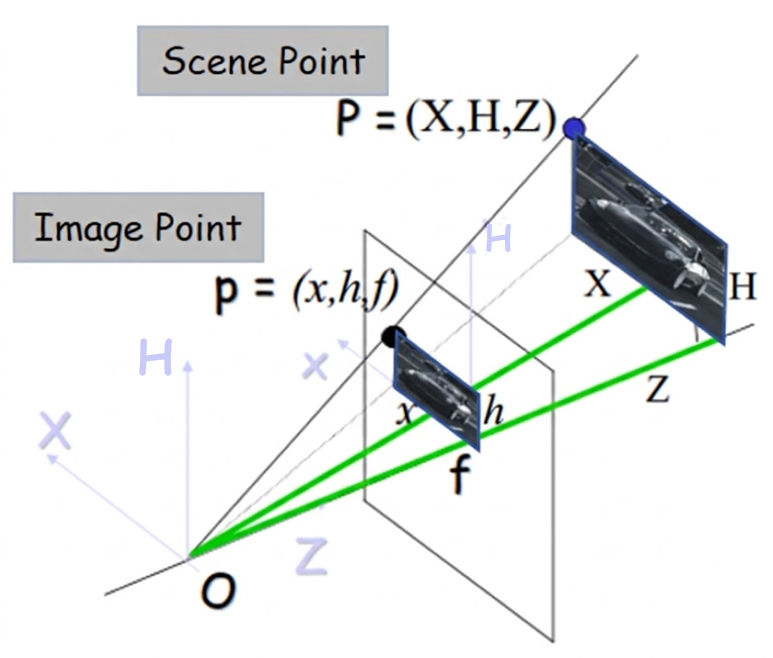}
        \caption{Camera Model}
        \label{fig:camera}
    \end{subfigure}
    \hfill % 使用 \hfill 填充中间空隙，确保两图对齐到左右两端
    % --- 右边：Point (子图 b) ---
    \begin{subfigure}[b]{0.68\linewidth}
        \centering
        \includegraphics[width=\linewidth]{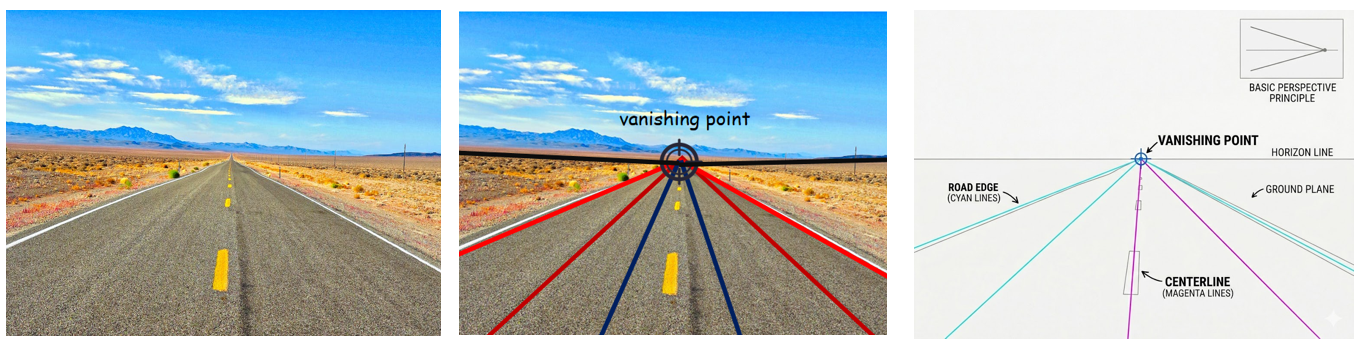}
        \caption{Perspective Convergence}
        \label{fig:VP}
    \end{subfigure}
    
    \caption{\textbf{Geometric principles of perspective.} 
    (a) The pinhole camera model mapping a 3D scene point $\mathbf{P}$ to a 2D image point X$\mathbf{p}$. 
    (b) Perspective convergence illustrating the vanishing point (VP). Left: Original sequence; Middle: Geometric overlay; Right: Schematic diagram of ground plane projection.}
    \label{fig:geo_perspective}
\end{figure}
To prove the identity in Eq.~\eqref{eq:hvp}, assume a 3D point $\mathbf{P}_t$ on a rigid object moving linearly in world space:
\[
\mathbf{P}_t = \mathbf{P}_0 + s_t\mathbf{d}, \quad \text{where } \mathbf{d}=(D_X, D_Y, D_Z) \text{ and } D_Z \neq 0.
\]
The pinhole projection $(x_t, y_t)$ is given by $x_t = f X_t / Z_t$ and $y_t = f Y_t / Z_t$. As the object recedes ($s_t \to \infty$), its image coordinates converge to the vanishing point:
\[
VP = \left( \lim_{s_t \to \infty} \frac{f(X_0 + s_t D_X)}{Z_0 + s_t D_Z}, \lim_{s_t \to \infty} \frac{f(Y_0 + s_t D_Y)}{Z_0 + s_t D_Z} \right) = \left( \frac{f D_X}{D_Z}, \frac{f D_Y}{D_Z} \right).
\]
The horizontal distance between the centroid and the $VP$ is:
\[
VP_x - x_t = \frac{f D_X}{D_Z} - \frac{f X_t}{Z_t} = \frac{f (D_X Z_t - D_Z X_t)}{D_Z Z_t}.
\]
Substituting $X_t = X_0 + s_t D_X$ and $Z_t = Z_0 + s_t D_Z$, the numerator simplifies to:
\[
D_X(Z_0 + s_t D_Z) - D_Z(X_0 + s_t D_X) = D_X Z_0 - D_Z X_0 = \text{Constant}.
\]
Thus, the 2D Euclidean distance to the $VP$ follows the relationship $\operatorname{Dist}(\mathbf{p}_t, VP) = \frac{f \cdot \mathcal{C}}{Z_t}$, where $\mathcal{C}$ is a constant determined by the initial geometry and motion direction. Given that a rigid body's projected height is $h_t = \frac{f H}{Z_t}$, it follows that:
\[
h_t \propto \frac{1}{Z_t} \quad \text{and} \quad \operatorname{Dist}(\mathbf{p}_t, VP) \propto \frac{1}{Z_t}.
\]
Dividing the initial state by the state at time $t$ yields the homogeneity relation:
\[
\frac{h_1}{h_t} = \frac{Z_t}{Z_1} = \frac{\operatorname{Dist}(\mathbf{p}_1, VP)}{\operatorname{Dist}(\mathbf{p}_t, VP)}.
\]
This concludes the derivation, proving that scale changes and radial convergence must be linearly coupled in 3D-consistent videos.

\section{Qualitative Visualizations}
\label{app:visual_details}

\subsection{Visualizing the PDI Computation Pipeline}
\label{app:pipeline_visualization}

To demonstrate the transparency, interpretability, and robust engineering of our proposed metric, we walk through the step-by-step execution of the PDI evaluation pipeline on a representative generated video (e.g., a vehicle driving forward). The pipeline systematically extracts multi-modal geometric evidence to compute the final physical realism score.

\begin{figure}[htbp]
    \centering
    \includegraphics[width=0.98\linewidth]{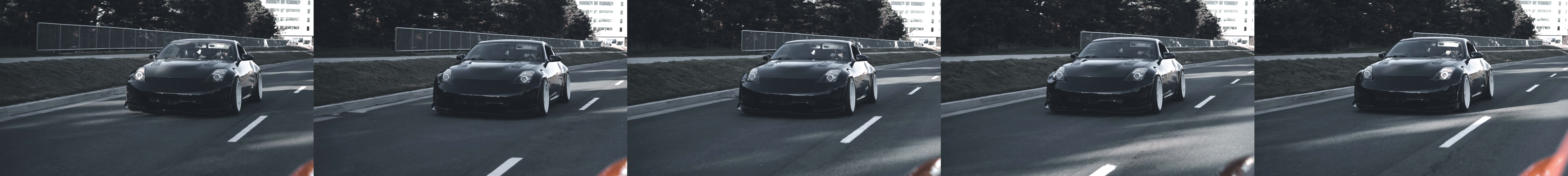}
    \caption{\textbf{Input Video Sequence.} Five uniformly sampled frames from a representative generated video featuring a moving vehicle. This raw temporal sequence serves as the initial input for our PDI computation pipeline.}
    \label{fig:input_sequence}
    \vspace{-10pt}
\end{figure}

\begin{figure}[htbp]
    \centering
    \begin{subfigure}{0.46\textwidth}
        \centering
        \includegraphics[width=\linewidth]{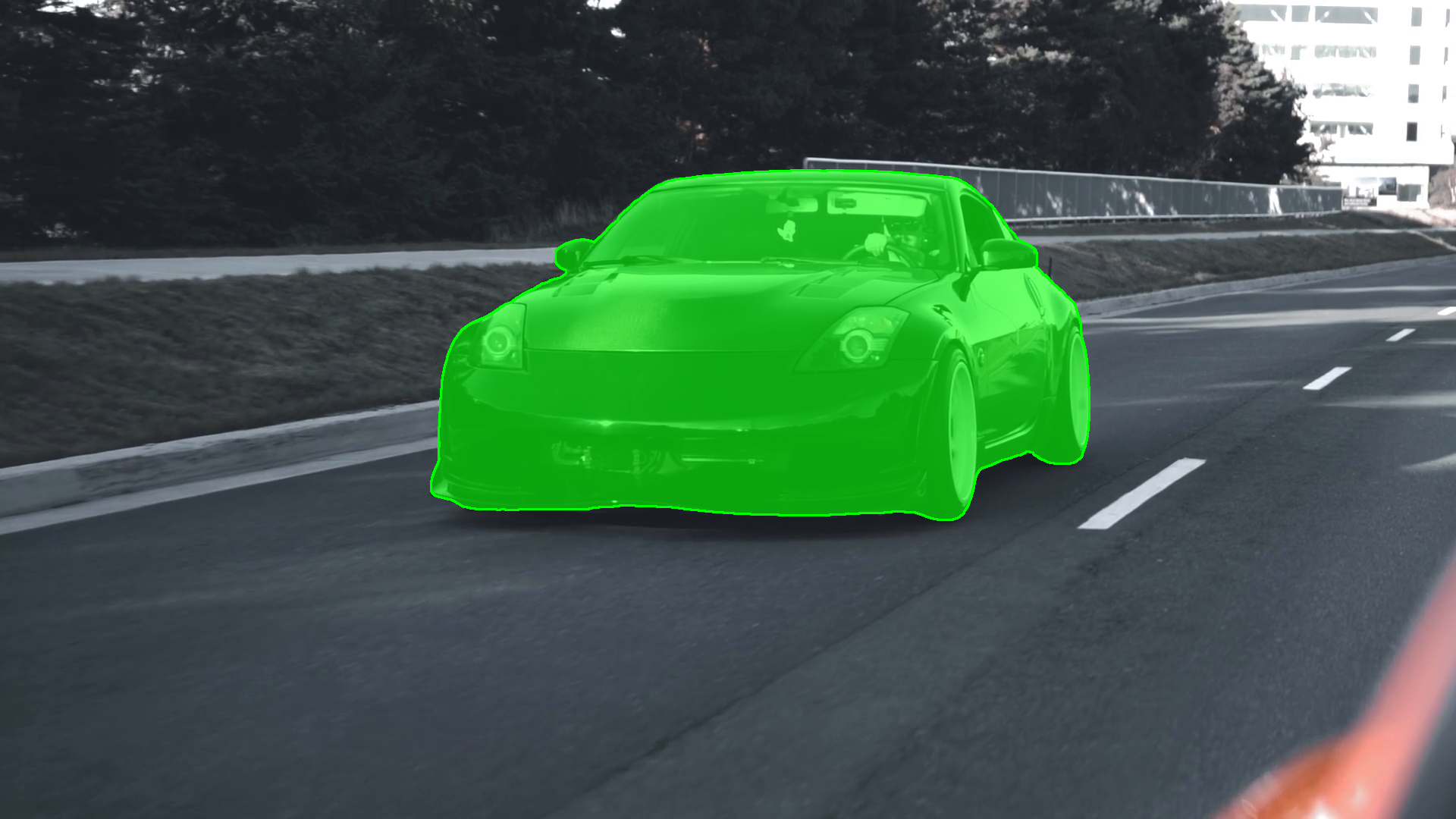}
        \caption{\textbf{Target Isolation.} SAM~2 accurately isolates the target vehicle.}
        \label{fig:pipe_mask}
    \end{subfigure}\hfill
    \begin{subfigure}{0.48\textwidth}
        \centering
        \includegraphics[width=\linewidth]{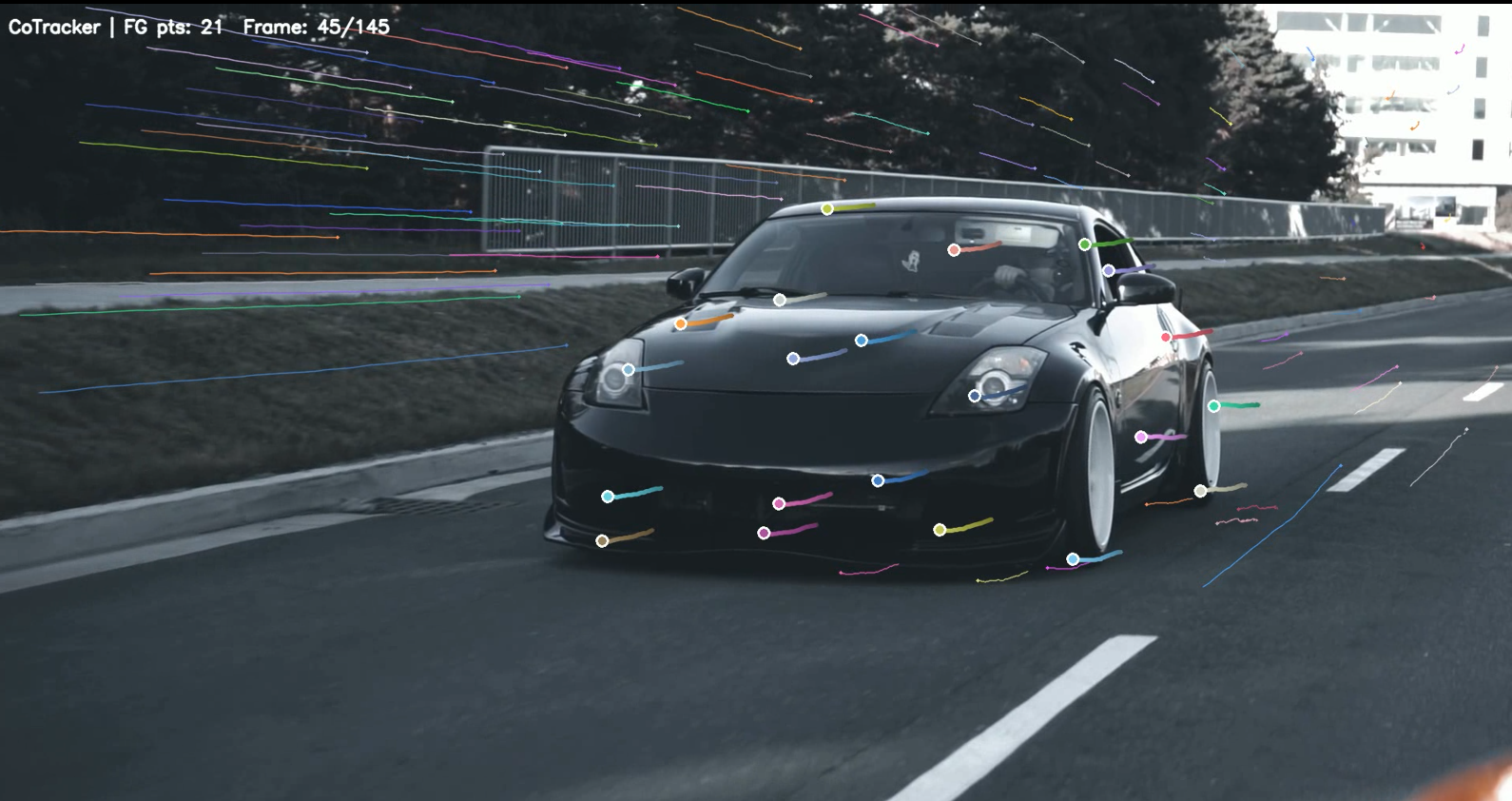}
        \caption{\textbf{Dense Tracking.} CoTracker3 produces long-term 2D trajectories inside the target mask.}
        \label{fig:pipe_track}
    \end{subfigure}
    \caption{\textbf{Pipeline intermediate 2D processing.} The system first segments the semantic target (a) and subsequently extracts dense kinematic trajectories (b) for downstream auditing.}
    \label{fig:pipeline_2d}
\end{figure}

\vspace{0.5em}
\noindent\textbf{Step 1: Semantic Targeting (SAM 2).}
We initiate the pipeline by identifying the auditing subject using \textbf{Florence-2} for automated text-to-box prompting. These prompts are then fed into \textbf{SAM~2} to generate and propagate a temporal sequence of binary masks $\{M_t\}_{t=1}^T$. This step isolates the subject and provides frame-wise instantaneous pixel heights $h_t$ and 2D spatial boundaries (Figure~\ref{fig:pipe_mask}).

\begin{figure}[htbp]
    \centering
    \begin{minipage}{0.32\linewidth}
        \centering
        \includegraphics[width=\linewidth]{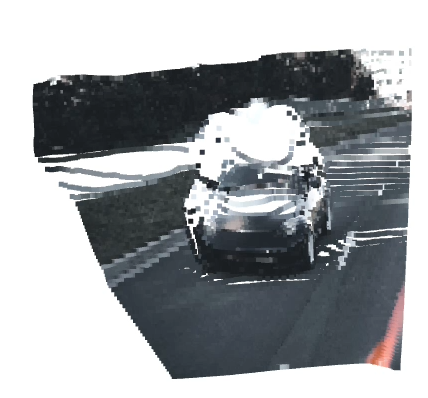}\\
        \small (a) Frame $t_1$
    \end{minipage}\hfill
    \begin{minipage}{0.32\linewidth}
        \centering
        \includegraphics[width=\linewidth]{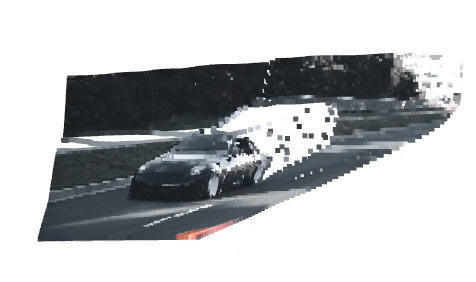}\\
        \small (b) Frame $t_2$
    \end{minipage}\hfill
    \begin{minipage}{0.32\linewidth}
        \centering
        \includegraphics[width=\linewidth]{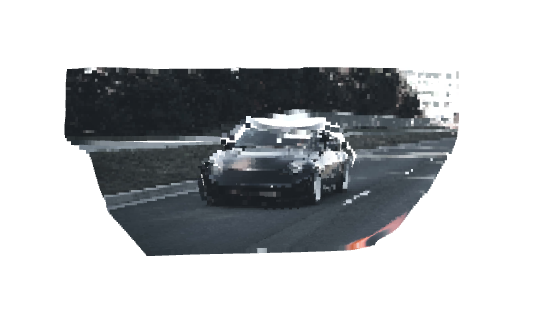}\\
        \small (c) Frame $t_3$
    \end{minipage}
    \vspace{4pt}
    \caption{\textbf{Dense 3D Reconstruction via MegaSaM.} Multi-view visualizations of the recovered 3D point clouds. Extracting these temporally consistent physical structures establishes a reliable geometric foundation for auditing spatial rigidity.}
    \label{fig:megasam_recon}
\end{figure}

\vspace{0.5em}
\noindent\textbf{Step 2: 3D Geometric Uplifting (MegaSaM).}
To recover the latent 3D physical environment, we employ \textbf{MegaSaM} to obtain a coherent depth sequence $\{Z_t\}_{t=1}^T$, the estimated focal length $f$, and camera poses. More importantly, MegaSaM projects every pixel into a unified 3D world coordinate system, yielding world-space pointmaps $\mathbf{P}_{world} \in \mathbb{R}^{T \times H \times W \times 3}$ (Figure~\ref{fig:megasam_recon}). This critical step lifts 2D observations into pure 3D space, decoupling object kinematics from camera ego-motion.

\vspace{0.5em}
\noindent\textbf{Step 3: 3D Structural Anchoring (CoTracker3).}
With the 3D world-space constructed, we deploy \textbf{CoTracker3} to monitor the subject's internal structural integrity. Within the region defined by $M_1$, we seed anchor queries and obtain a set of reliable 2D pixel-space trajectories $\{(u_t^n, v_t^n)\}$ (Figure~\ref{fig:pipe_track}). By using these as spatial indices into the MegaSaM pointmaps, we lift each tracked anchor to its 3D coordinate: $\mathbf{q}_t^n = \mathbf{P}_{world}[t, v_t^n, u_t^n]$, transforming 2D visual tracking into structurally meaningful 3D trajectories.

\vspace{0.5em}
\noindent\textbf{Step 4: Three-Dimensional Geometric Auditing.}
Utilizing the extracted 2D tracks and 3D geometries, the system audits the object's physical rationale across three orthogonal dimensions:
\begin{itemize}
    \item \textbf{Scale ($\epsilon_{scale}$):} Audits whether the target's projected height and depth satisfy the $h \cdot Z = \text{const}$ invariant (pinhole camera model).
    \item \textbf{Trajectory ($\epsilon_{traj}$):} Evaluates whether the centroid motion in 3D world coordinates follows Newtonian inertia, penalizing abrupt acceleration jumps or non-physical reversals.
    \item \textbf{Rigidity ($\epsilon_{rigidity}$):} Quantifies the temporal instability of the 3D distances between anchor pairs via the Coefficient of Variation, penalizing ``volume breathing'' or ``Jello effect'' artifacts.
\end{itemize}
\vspace{0.5em}
\noindent\textbf{Step 5: Score Synthesis \& Final Report.}
The individual dimension errors are aggregated into the final Perspective Distortion Index (PDI) using the weights $(w_1=0.4, w_2=0.4, w_3=0.2)$:
\begin{equation}
    \text{PDI} = w_1 \cdot \text{RMSE}(\epsilon_{scale}) + w_2 \cdot \text{RMSE}(\epsilon_{traj}) + w_3 \cdot \epsilon_{rigidity}
\end{equation}
A lower PDI score indicates superior physical rationale. The system ultimately generates a terminal audit report (shown below), providing fine-grained diagnostics to pinpoint specific failure modes.

\section{Qualitative Analysis of Failure Modes}
\label{app:failure_cases}

While PDI-Bench provides robust quantitative scores, qualitative examination reveals distinct geometric failure modes that persist even in state-of-the-art generative models. 

\begin{figure}[h]
    \centering
    \begin{subfigure}[b]{\linewidth}
        \includegraphics[width=\linewidth]{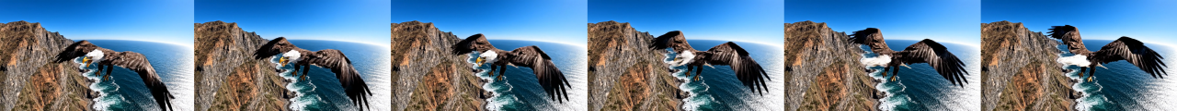}
        \caption{Biological Motion: Articulated dynamics under rapid motion.}
    \end{subfigure}
    \begin{subfigure}[b]{\linewidth}
        \vspace{5pt}
        \includegraphics[width=\linewidth]{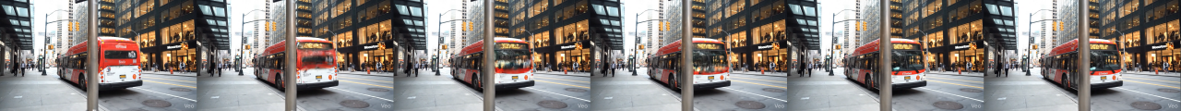}
        \caption{Longitudinal Convergence: Geometric artifacts during camera-object interaction.}
    \end{subfigure}
    \caption{\textbf{Qualitative Failure Modes.} (a) A bird in flight exhibiting motion-induced blur and feature morphing. (b) A bus sequence showing temporal inconsistencies where the object's appearance and 3D structure become spatially decoupled across consecutive frames.}
    \label{fig:flip_failure}
\end{figure}

\paragraph{Temporal Orientation and Structural Incoherence.} 
A unique failure mode observed in models like \textbf{Veo 3.1-Fast} is the abrupt inversion or morphing of object features, where frontal and rear characteristics appear to swap or deform mid-sequence (Fig.~\ref{fig:flip_failure}). Crucially, these artifacts often manifest as \textit{instantaneous events}. As the transition occurs within a duration shorter than the temporal window of our SfM-based reconstruction, the underlying pointmaps struggle to track the rapid topology change, leading to transient spikes in \textit{Structural Rigidity} ($\epsilon_{r}$) and \textit{Motion Consistency} ($\epsilon_{t}$). While PDI-Bench captures the resulting geometric inconsistency, these extreme behaviors highlight the difficulty of monocular 3D systems in maintaining a coherent object-centric frame of reference during non-physical transitions.

\section{Limitations and Future Directions}
\label{app:limits}

While PDI-Bench establishes a rigorous geometric yardstick, several limitations remain.
First, the framework inherits a dependency on off-the-shelf perception modules: when SAM~2, CoTracker3, or MegaSAM fail in low-texture or low-parallax regimes, the system must fall back to 2D proxies and reconstruction-aware weighting, reducing the depth sensitivity of the final score.
Second, our core invariants are derived under a rigid-body assumption; although the rigidity residual partially compensates for mild non-rigid motion, highly deformable or amorphous objects (e.g., fluids, cloth, or crowds) are not fully captured by the current formulation.
Third, disentangling complex 3D rotations from axial translation using monocular cues is fundamentally ill-posed; the Model Consensus mechanism mitigates most false positives, but residual noise can persist in extreme curved-motion scenarios.

We see three main avenues for future work.
One direction is to integrate learned priors by distilling PDI-Bench into a lightweight neural assessor that approximates our geometric residuals without explicit SfM, enabling larger-scale or near real-time auditing.
Another is to extend the current Euclidean invariants to cover richer physical phenomena, such as fluid dynamics, articulated bodies, and multi-object interactions, potentially via hybrid Lagrangian--Eulerian formulations.
Finally, incorporating multi-view or multi-sensor inputs (e.g., stereo, depth sensors, IMU traces) could reduce the inherent ambiguities of monocular geometry and allow PDI-Eval to serve as a calibration tool for future, more physically grounded world models.

\section{Broader Impacts}
\label{app:broaderimpacts}
This work introduces PDI-Bench, a benchmark designed to evaluate the physical consistency and geometric realism of video generation models. We believe the proposed benchmark can provide several positive societal impacts. First, by enabling systematic evaluation of geometric and physical plausibility, PDI-Bench may help improve the reliability and interpretability of generative video systems. Such evaluation tools can benefit applications in robotics, embodied AI, simulation, digital content creation, and educational visualization, where physically coherent video generation is important. Second, the benchmark promotes transparency by exposing failure modes of current video generation models, which may encourage the development of safer and more trustworthy generative systems.

At the same time, advances in realistic video generation may also introduce potential risks. Improving the physical realism of generated videos could contribute to the creation of increasingly convincing synthetic media, which may be misused for misinformation, deceptive content generation, or malicious manipulation. In addition, highly realistic generated videos could complicate the detection of synthetic media in sensitive contexts.

By quantitatively measuring physical consistency and identifying model weaknesses, the benchmark may support future research on trustworthy generative modeling, robustness analysis, and synthetic media detection. We encourage future work to further investigate safeguards and responsible deployment practices for realistic video generation technologies.

\clearpage

\end{document}